%% file: main.tex
\documentclass[runningheads]{llncs}

\usepackage{eccv}

\usepackage{eccvabbrv}

\input{preamble}

\usepackage[accsupp]{axessibility}  % Improves PDF readability for those with disabilities.

\usepackage[pagebackref,breaklinks,colorlinks,allcolors=eccvblue]{hyperref}

\usepackage{orcidlink}

\begin{document}

% ---------------------------------------------------------------
% TODO REVIEW: Replace with your title
\title{ReDesign: Recovering Editable Design Structures from Images via Agentic Decomposition} 
% ReDesign: Recovering Editable Design Structures from Raster Images via Agentic Decomposition

% TODO REVIEW: If the paper title is too long for the running head, you can set
% an abbreviated paper title here. If not, comment out.
\titlerunning{ReDesign}

% TODO FINAL: Replace with your author list. 
% Include the authors' OCRID for the camera-ready version, if at all possible.
\author{
Jooyeol Yun*\inst{1}\and
Jintae Park*\inst{1}\and
Hyesu Lim\inst{2}\and
Junha Hyung\inst{1}\and
Hyungjin Chung\inst{3} \inst{4} \and
Jaegul Choo\inst{1}
}

% TODO FINAL: Replace with an abbreviated list of authors.
\authorrunning{Yun et al.}
% First names are abbreviated in the running head.
% If there are more than two authors, 'et al.' is used.

% TODO FINAL: Replace with your institution list.
\institute{KAIST AI, Daejeon, Korea \\
\email{\{blizzard072, sonjt00 , sharpeeee, jchoo\}@kaist.ac.kr}\and
Helmholtz Munich, Neuherberg, Germany \\
\email{hye.lim@helmholtz-munich.de} \and
Korea University, EverEx\\
\email{\ hj.chung@korea.ac.kr}
}

\maketitle
\renewcommand{\thefootnote}{\fnsymbol{footnote}}
\footnotetext[1]{\vspace{-2.5cm} indicates equal contribution.}
\renewcommand{\thefootnote}{\arabic{footnote}}

\vspace{-0.5cm}
\begin{figure*}[h]
    \centering
    \includegraphics[width=\linewidth]{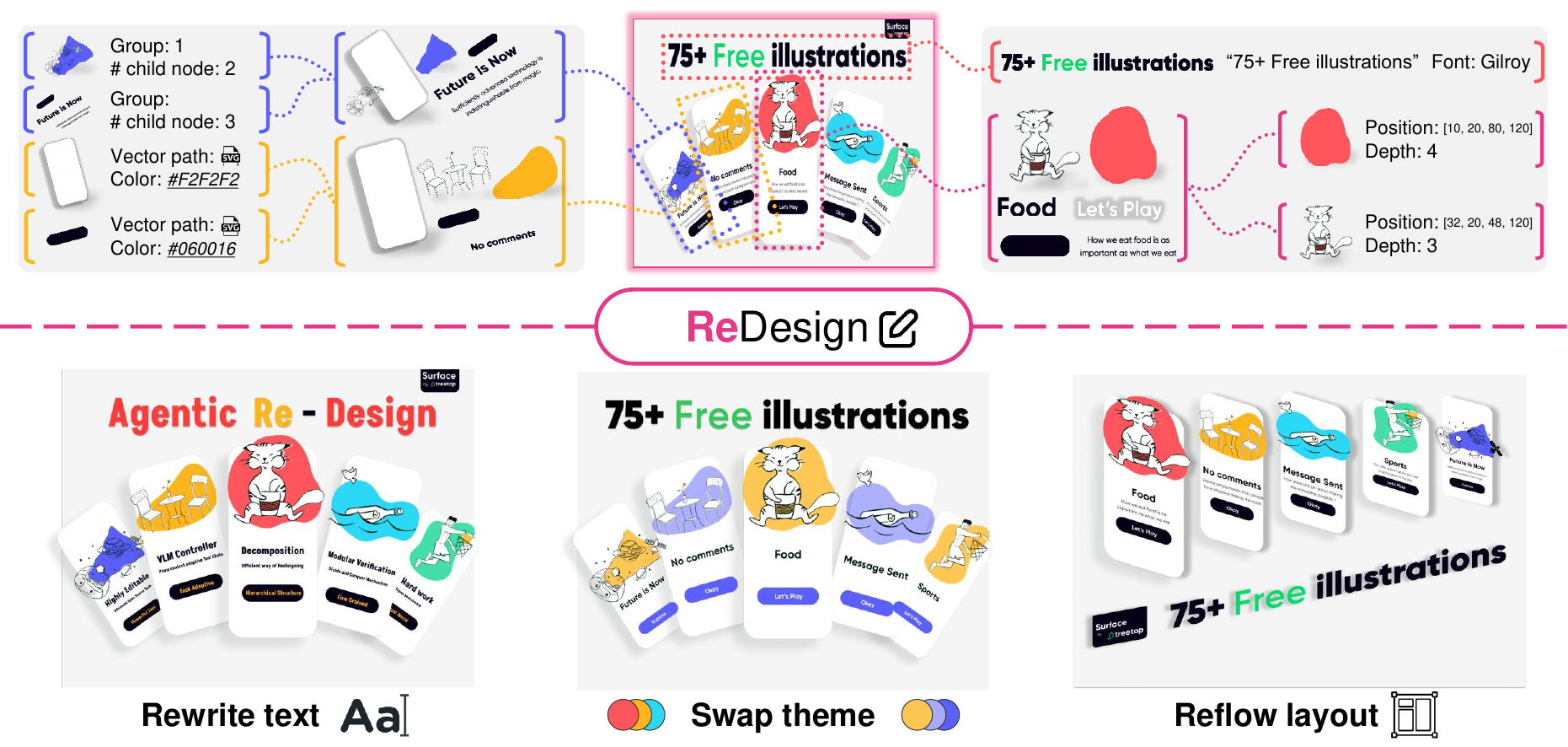}
    \caption{ReDesign turns raster images into editable representations that enable real-time adjustments, such as rewriting text, color theme swaps, and repositioning layouts, directly to designs. Visit the \href{https://jintae-00.github.io/ReDesign/}{project page} for interactive demos.
    }
    \vspace{-1cm}
\end{figure*}

\begin{abstract}
Recovering an editable design file from a raster image is a common and costly bottleneck in modern design workflows, yet remains challenging since editability depends on recovering multi-modal attributes, such as typography, vector geometry, colors, grouping, and layer ordering. 
We present ReDesign, an agentic framework that grows an editable layer hierarchy by selecting and composing specialized tools across modalities. 
To keep this long decision process reliable despite imperfect tool outputs, we introduce graceful verification at each expansion, which provides local accept, prune, or retry feedback that prevents error accumulation and avoids large scale reruns.
To evaluate editability at scale, we introduce the Figma Edit Replay Benchmark, consisting of 909 raw Figma files and 14,796 controlled edit instructions that replay edits on reconstructed outputs. 
Across this benchmark and standard reconstruction metrics, ReDesign achieves strong visual fidelity while delivering the highest editability across layout, color, and text edits, outperforming layered decomposition baselines and serial tool use pipelines.
% \keywords{Design understanding \and Vision language models}
\end{abstract}

\section{Introduction}
\label{sec:introduction}
% What is our problem? Why is it important?
Designers routinely need to adapt a single design across media platforms, repurpose layouts for new campaigns, and make adjustments for accessibility. 
In practice, many real world design assets and handoff deliverables exist only as raster exports or screenshots, so the preceding task is to reconstruct an editable version (\eg, Figma or Adobe Illustrator file) from a raster image. 
% In practice, old design files are often lost, so the preceding task is to reconstruct an editable version (\eg, ???) from an exported raster image or screenshot. 
Today, that reconstruction is largely manual and error prone, forcing designers into tedious recreation before any meaningful edits can begin. 
Thus, a system that automatically recovers editable components such as text layers, vector shapes, and layouts can remove this tedious step and dramatically reduce the turnaround time for designers. 

% Why is it hard?
Making a raster image re-editable means recovering the same structured representation that designers work with. 
Design tools, such as Adobe Illustrator, Figma, and Canva,  keeps typography, color, and layout as separate objects with their own parameters because that aligns with how humans perceive and edit a design. 
This structure is exactly what makes editing easy, as these representations expose the correct ``knobs'', such as font attributes for text, vector paths for primitive shapes, and position and scale for layouts. 
In essence, rasterization is a many-to-one mapping from editable layer structures to pixels, so recovering the original structure from an image is an ill-posed inverse problem. 
% Reversing that process is hard for the same reason editing is easy. 
While the final appearance may look identical, a raster export no longer states which pixels belong to which object, which attributes produced them, or how elements were layered. 
Therefore, a system has to infer a plausible set of objects and parameters across multiple modalities including text, geometry, color, and layout, so the result both matches the image and behaves like an editable design.

% What did existing studies do?
% What is our solution
Many previous studies solve key subproblems for editable reconstruction, such as recognizing text~\cite{ocr}, decomposing layers~\cite{layerd, qwen-image-layered,omnipsd}, and extracting shapes~\cite{omnisvg,starvector}. 
A natural next step is to combine these components into an end-to-end system, but a straightforward composition is often unreliable, since errors from one tool can cascade and a single fixed sequence rarely fits the diversity of real designs. 
In this paper, we present ReDesign, which addresses this gap by focusing on the structure of the editable output. 
Editable designs naturally come as a layer hierarchy, with groups and layers organized in a tree, and we therefore cast reconstruction as recovering this hierarchy from a raster image. 
Starting from the full image as the root, ReDesign operates as a vision-language model (VLM) agent that repeatedly selects a node to expand and composes a tool sequence from a fixed action set to produce child nodes, progressively refining the hierarchy into atomic editable elements.

% Example figure where tool outputs fail?

In practice, tool outputs are often imperfect, and the controller can make incorrect expansion choices, producing branches that drift away from the actual structure of the original design. 
An accept or reject decision on the final output~\cite{shinn2023reflexion,chen2025rethinking} may be easy to compute but offers little guidance about which expansion failed or how to repair it. 
We therefore introduce a graceful verification at every node expansion, where a verifier evaluates the proposed children, and either accepts the expansion, prunes invalid branches, or retries with a different tool or configuration. 
The verifier targets failure modes specific to hierarchical growth, such as duplication across sibling nodes and incomplete coverage of the parent region.
This keeps tree growth efficient by pruning and retrying only the offending branch at the moment it diverges, rather than letting errors accumulate until they force a hard failure.

To evaluate our method, we introduce the Figma Edit Replay Benchmark, a collection of diverse real world designs from raw Figma files~\cite{figma} with ground-truth layer hierarchies and attributes. 
Using this benchmark, we evaluate both visual fidelity and editability, with editability measured by replaying 14,796 controlled edit instructions spanning layout, color, and text. 
Our evaluations show that ReDesign achieves superior performance compared to strong baselines.

Our contributions are threefold:
% \begin{itemize}
%   \item[$\bullet$] We cast raster-to-editable reconstruction as a structured tree expansion problem, and build an agentic pipeline that grows an editable layer hierarchy into atomic elements.
%   \item[$\bullet$] We introduce graceful verifications at each node expansion, providing step-level feedback for local revision and preventing cascading failures from imperfect tool outputs.
%   \item[$\bullet$] Our extensive experiments on our editability benchmark of diverse real world designs paired with raw Figma files, reveal that our approach preserves visual fidelity and editability with supporting ablations and comparisons against strong baselines.
% \end{itemize}
Our contributions are threefold:
\begin{itemize}
  \item[$\bullet$] We introduce ReDesign, an agentic raster-to-editable reconstruction system that casts the problem as structured tree expansion and grows an editable layer hierarchy into atomic elements through tool composition.
  \item[$\bullet$] The graceful verification enforces local correctness at each node expansion with accept, prune, and retry outcomes, enabling targeted repair and preventing hard failures from cascaded tool errors.
  \item[$\bullet$] We construct the Figma Edit Replay Benchmark and conduct extensive experiments showing that ReDesign achieves strong visual fidelity and state-of-the-art editability across layout, color, and text edits, with supporting ablations and comparisons against strong baselines.
\end{itemize}

\section{Related Work}
\label{sec:related_work}

% Layered image generation (+ Nano banana and Flux-Kontext)
\subsection{Layered Image Generation and Decomposition}
A growing line of work studies layer-aware image generation~\cite{fontanella2024generating,pu2025art,liu2020learning,kang2025layeringdiff,chen2025prismlayers,chen2025posta,zhang2025creatiposter} and decomposition~\cite{wang2025diffdecompose,layerd}, motivated by the need for controllable composition and post-hoc editing~\cite{zhang2023text2layer,huang2024layerdiff,zhang2024transparent,chen2025rethinking}.
Closest to our setting are methods that decompose raster graphics into layers or generate PSD-like layered outputs, aiming for inherent editability~\cite{layerd,qwen-image-layered,omnipsd}. 
These approaches provide useful primitives for isolating elements and for constraining edits to localized regions, and they become even more practical when combined with modern image editing models~\cite{nanobanana,fluxkontext}.
However, producing layers is only part of reconstructing an editable design file, which also requires correct semantics, hierarchy, and parameters so that edits behave as expected.
We therefore treat layered decomposition models as tool for a larger decomposition workflow, rather than as a complete solution. 

\subsection{Image Vectorization and SVG Generation}
Image vectorization has a long history, from classical tracing pipelines~\cite{vtracer} to learned models that generate vector primitives or scalable vector graphics (SVG) programs~\cite{deepsvg,starvector,omnisvg,layertracer}.
Vectorization is an important ingredient for reconstructing editable shapes and icons, since vector paths expose direct controls for geometry, scale, and appearance. 
At the same time, vectorization is not universally desirable, since forcing complex textures or photographic content into paths can reduce editability and distort the representation relative to how the design was authored.
Practical reconstruction therefore requires deciding when vector paths are appropriate, and when regions should instead be represented as text or raster layers.
Thus, similarly, we utilize vectorization methods as one set of tools within our hierarchical decomposition process, selectively applying them where they best support downstream editing.

\subsection{Tool-Using Agents}
Tool-using agents extend language and vision-language models~\cite{gemini,gpt5} with external modules, enabling multi-step problem solving through action selection and execution~\cite{yao2022react,schick2023toolformer}.
In parallel, modern image editing models provide practical primitives for localized manipulation, content removal, and background completion that can support iterative pipelines~\cite{fluxkontext,nanobanana}.
Most existing agentic pipelines for vision emphasize serial tool invocation, often paired with an end-stage validation of the final output~\cite{shinn2023reflexion,qi2024mutual}.
Editable reconstruction differs in that it requires a long sequence of structural decisions, where early commitments change the state of the reconstruction and constrain what later tools can recover.
This makes reliability depend on how intermediate decisions are guided and revised during the process, rather than only on the strength of individual tools or a final verification step.
Our work builds on these agentic and editing primitives, and focuses on structured sequential decision making that supports local revision while constructing an editable layer hierarchy.

\section{Method}

\subsection{Problem Setup and Overview}
\begin{figure}[t!]
    \centering
    \includegraphics[width=1.0\linewidth]{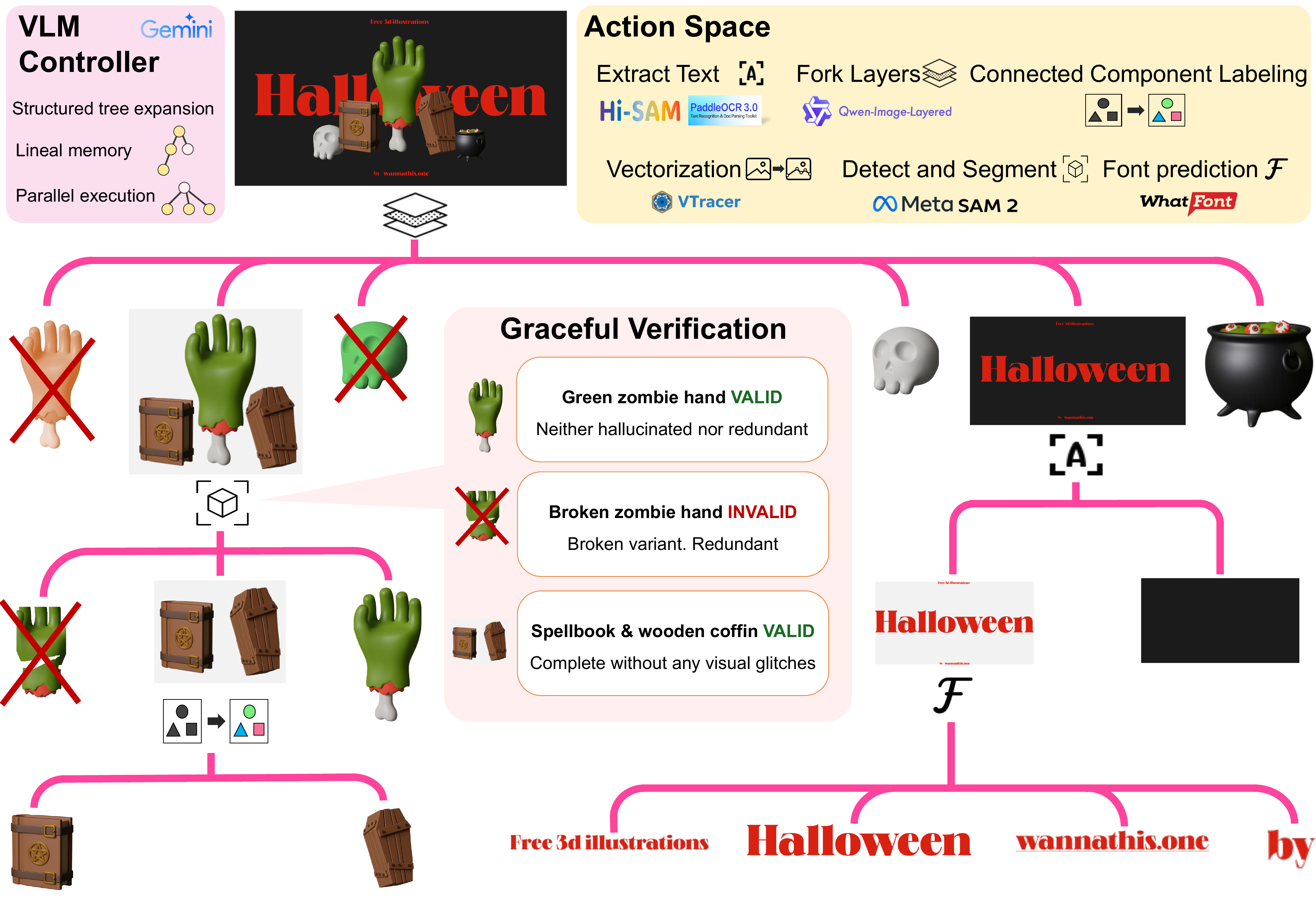}
    \caption{Overview of ReDesign. A VLM controller grows an editable layer hierarchy by selecting from a fixed set of tool actions for text, objects, and layered decomposition, and runs graceful verification at each parent-to-children expansion to accept valid splits, prune redundant or hallucinated children, or trigger a retry, until atomic editable elements are recovered.}
    \label{fig:method-main}
    % \vspace{-1cm} %
\end{figure}

Given an input raster image, our goal is to recover an editable design structure, as we depict in \Cref{fig:method-main}.
The desired output is a JSON hierarchy that has editable meta-data, such as vector shapes, colors, text contents, fonts, groups, and z-orders, and this tree hierarchy must be inferred from the root node which is the raster image.

We maintain a partial reconstruction tree, which is a tree that contains nodes whose metadata fields may be incomplete and are filled in over time. 
The partial reconstruction tree is expanded as the controller policy selects an action from a fixed set of actions. 
A graceful verifier evaluates the proposed expansion at every step and either accepts the proposed children nodes the or provides signals to revise the actions.
The process terminates when the tree has sufficient metadata to be exported as an editable JSON hierarchy.

\subsection{Action Space and Tools}
Converting a raster design into an editable format requires handling heterogeneous element types, such as text, vector shapes, photographs, and layered compositions, each demanding different specialized processing.
We therefore define a discrete action space in which each action encapsulates a tool chain tailored to a specific decomposition scenario. 
This space is modular by design and can be extended as new tools become available.

\paragraph{\textup{\textbf{Text extraction.}}}
An optical character recognition tool~\cite{ocr} localizes and reads visible text, after which the text pixels are segmented~\cite{hisam} and the vacated region is inpainted~\cite{lama}, producing a text layer and a remainder layer with the remainder background.

\paragraph{\textup{\textbf{Multi-layer decomposition.}}}
A layered image generation model~\cite{qwen-image-layered} offers powerful decomposition capability separating the raster image into distinct layers, with an adaptively chosen layer count based on visual complexity.

\paragraph{\textup{\textbf{Connected Component Labeling.}}}
When a layer contains multiple elements that are spatially disjoint, it can be split into non-overlapping children using a simple connected component labeling algorithm~\cite{cca}. 

\paragraph{\textup{\textbf{Detection and segmentation.}}}
For entangled objects, a VLM~\cite{gemini-3-flash} identifies the foremost element, an open-set detector~\cite{gdino} localizes it, a segmentation model~\cite{sam} extracts it, and an inpainting model~\cite{lama,objectclear} fills the cropped out region, yielding a foreground node and a background node.

\paragraph{\textup{\textbf{Vectorization.}}}
Leaf nodes are finalized as either vectorized SVG paths~\cite{vtracer} for shape-like content, or raster layers with placement metadata for photographic content.
Text leaves undergo font recognition~\cite{whatfontapi} and typographic property fitting to produce fully editable text elements carrying necessary font attributes.

\subsection{Controller Policy for Tree Expansion}
The core idea of the controller policy is to follow the same structure as an editable design file, instead of a serial or unstructured sequence of actions (tool calls). 
We therefore represent the reconstruction as a tree that is grown progressively from the full canvas into smaller, more specific regions, so early decisions establish high level grouping and ordering, and later decisions refine each group into editable leaves.
Each node represents a localized region together with its current metadata, and any attributes already inferred by previous tools.  
At each step, the controller selects an action given the node's partial rendering, and the expansion history stored for that node, which records prior tool choices and verification outcomes.  

Furthermore, since an expansion only depends on the parent node and its lineal state, and not on sibling states, our structured tree growth naturally enables parallel expansion across all current leaf nodes. 
Thus, we can expand multiple leaf nodes simultaneously, which improves throughput without changing the decision logic or sacrificing the coarse-to-fine consistency of the recovered hierarchy.

\subsection{Graceful Verification}
Structured expansion alone does not guarantee a correct reconstruction due to imperfect tools and occasional controller mistakes that propose duplicate content across siblings, leave parts of the parent region unexplained, or introduce content that was never present in the input. 
These errors are especially harmful since they immediately change the state of the tree and therefore constrain all later actions, resulting in a hard failure, where a system would have to re-run from the begining.

However, our structured expansion effectively limits the scope of each error to a single parent-to-children proposal, and therefore makes verification both local and well-defined, as the goal of an expansion is simply to decompose one parent region into a set of children that jointly explain it. 
Thus, we attach a graceful verification step to every proposed expansion and evaluate whether the candidate children form a valid decomposition under two simple criteria, (1) does the union of child nodes cover the parent content and (2) does the child node expand beyond the parent content. 
The verifier either accepts the expansion, prunes invalid or redundant children, or triggers a retry by requesting a different tool choice or adjusted hyperparameters for the same parent. 
In this way, errors are corrected locally before deeper branches are built on top of them, and the controller receives structured feedback tied to a single parent to children decision, which in turn reduces hard failures that would require large-scale reruns.

\subsection{Memory Management and Repair Signals}
We maintain a \emph{lineal} memory per node, meaning that each expansion stores only its own history along the path from the root rather than tracking sibling states, and this design avoids exponential growth while also preventing concurrency issues due to  parallel execution. 
Specifically, for every node we record the tool inputs and outputs, the tool choice and hyperparameters, and any verification outcomes or failure reasons, and this compact record makes the controller's next decision better grounded because it can reuse what worked and avoid repeating configurations that already failed. 
Therefore, when an expansion is rejected or pruned by the verifier, the controller can perform local repair by selecting a different action or adjusted settings based on the stored failure signal, and continuing growth without disturbing the rest of the tree, which reduces redundant computation and keeps parallel progress stable.

\begin{figure}[t]
    \centering
    \includegraphics[width=\linewidth]{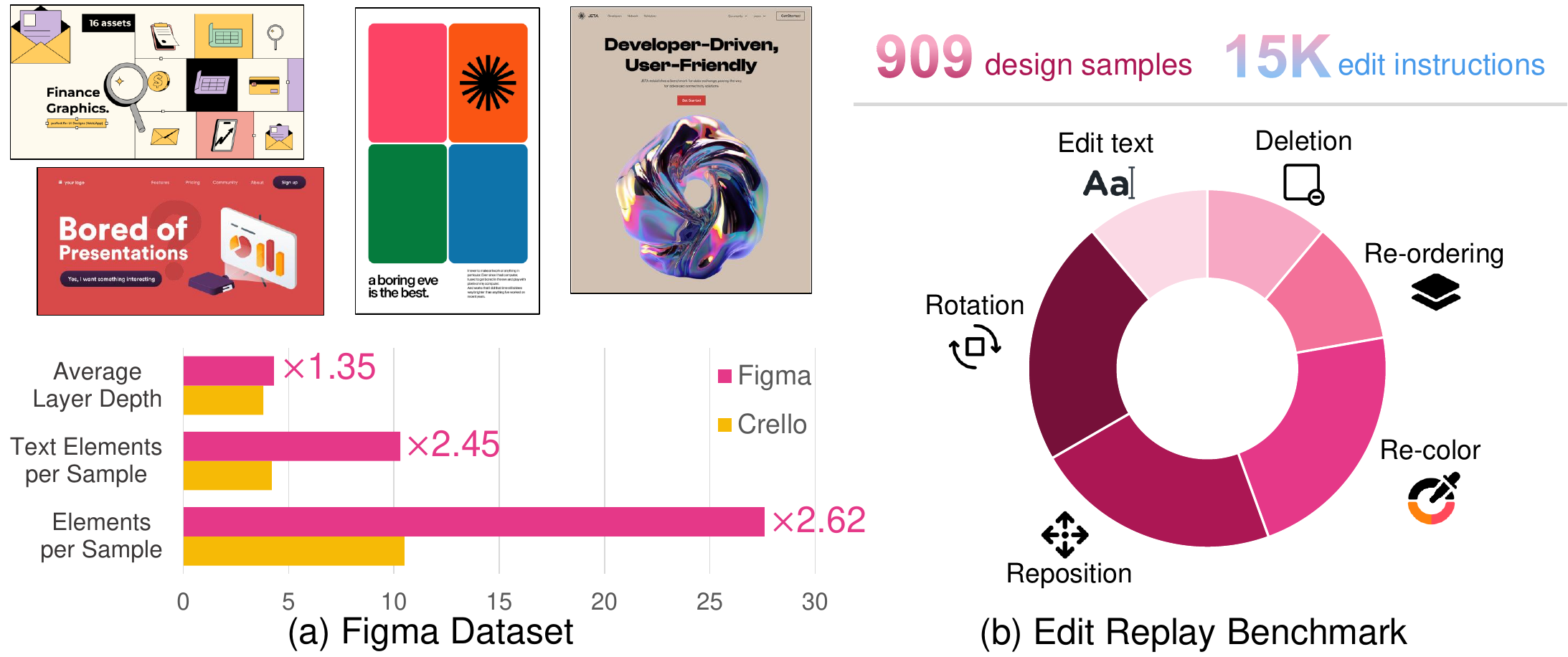}
    \caption{(a) Figma examples and statistics compared to the Crello dataset. (b) Edit Replay Benchmark and detailed edit types. }
    \label{fig:exp-dataset}
    \vspace{-0.3cm}
\end{figure}

\section{Experiments}
\subsection{Figma Dataset and the Edit Replay Benchmark}
To evaluate reconstruction on real-world design artifacts with ground-truth structure, we collect a dataset of 909 raw Figma design files. 
Each file provides the layer hierarchy along with element types and attributes, including text content and typography, vector shapes, colors, and ordering. 
As seen in \Cref{fig:exp-dataset} (a), we find that the Figma dataset consists of complex real-world designs compared to the original Crello dataset~\cite{crello}, with an average 2.62 times more elements per sample. 

We use these files to benchmark editability by generating paired supervision. 
Specifically, for each design, we apply a controlled edit directly to the original Figma document (\eg, reposition, text edit, and recolor) and render the post-edit image as ground truth. 
In total, we curate 14,796 edit instructions, averaging 15 edits per design. 

We also evaluate visual reconstruction accuracy on the Crello dataset~\cite{crello}, which provides raster design images with annotations. 
Following standard practice, we report reconstruction metrics on Crello to enable comparison with prior methods that emphasize visual fidelity and element recovery.

\subsection{Baselines}
\paragraph{\textup{\textbf{Layered image decomposition.}}}
We compare against Qwen-Image-Layered~\cite{qwen-image-layered} and LayerD~\cite{layerd}, which decompose an input image into a set of RGBA layers.
Since RGBA images are not directly editable, we post process the layers using OCR and vectorization tools, as discussed in the original papers. 

\paragraph{\textup{\textbf{Tool using agent. }}}
We compare against a standard tool using agent, similar to a ReAct framework~\cite{yao2022react}, that has access to the same tools and memory and uses the same VLM backbone. 
The agent produces an editable JSON output through a single linear sequence of tool calls, with validation at the end. 

\paragraph{\textup{\textbf{Image vectorization. }}}
We compare against VTracer~\cite{vtracer}, a widely used image vectorization tool, and treat their outputs as an editable design.
We also experimented with recent vectorization models (\eg, OmniSVG~\cite{omnisvg} and StarVector~\cite{starvector}), but found that they perform poorly on design images due to a domain gap, since many are trained primarily on icons or simplified graphics.

\begin{figure}[t]
    \centering
    \includegraphics[width=\linewidth]{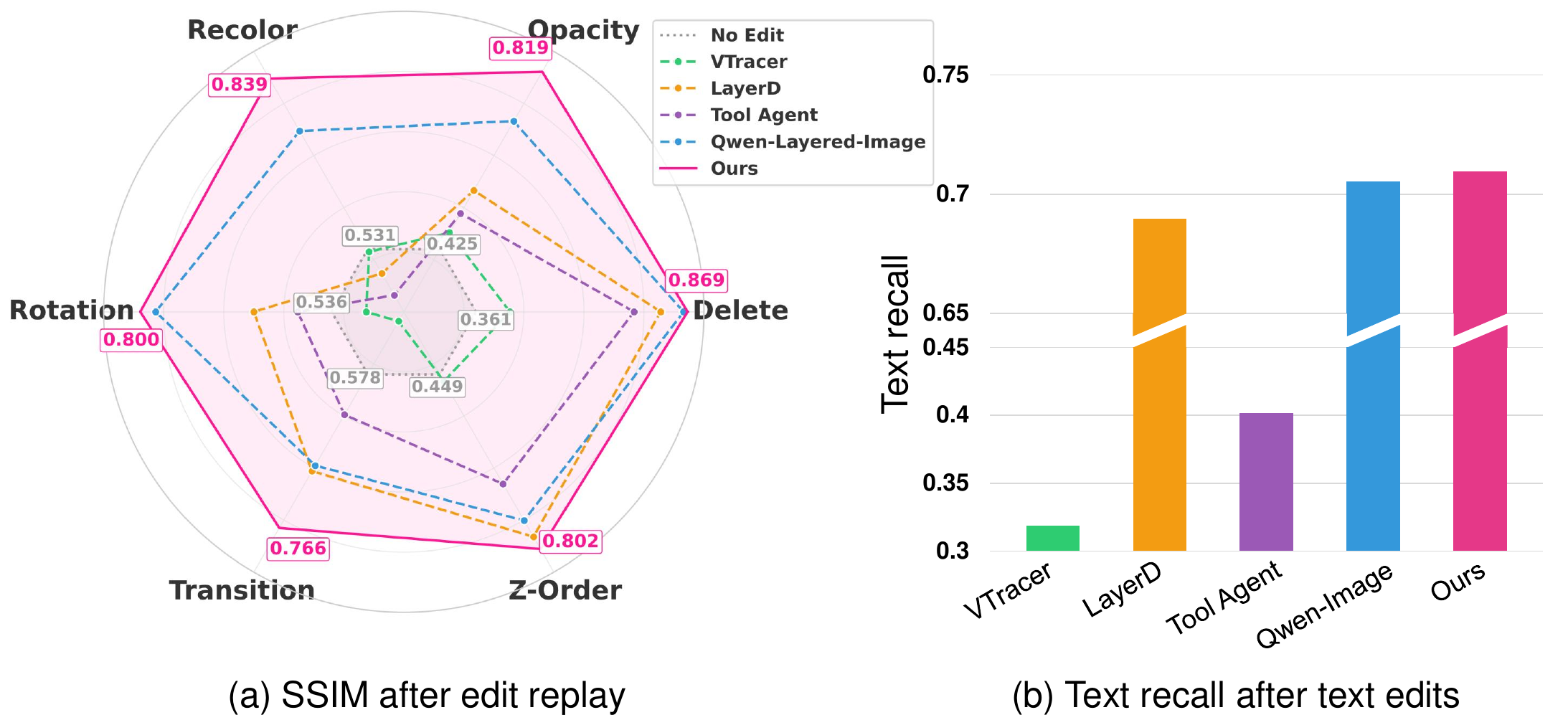}
    \vspace{-0.4cm}
    \caption{(a) Edit replay SSIM for different edit types, where our method performs best across all edits. (b) Text editability measured by text recall after replaying text edits, where our method achieves the highest recall among baselines.}
    \label{fig:exp-editability}
\end{figure}

\subsection{Editability}
\paragraph{\textup{\textbf{Edit replay protocol. }}}
We evaluate editability on the Figma design dataset using an edit replay protocol, meaning that we apply the same edit operation to the ground-truth design and to the reconstructed design, and then compare the rendered results. 
Each example provides a ground-truth editable design file and its corresponding raster image, after which each method takes the raster image as input and produces an editable output in the same target format. 
We then replay the edit instruction on the ground truth file to obtain a rendered reference, replay the same instruction on the predicted editable output, and finally measure agreement between the two edited renderings.

Specifically, to apply the same instruction to a predicted editable output, we first match the target ground-truth element to an element in the prediction. 
Since predictions made by different models have different representations of a reconstructed design, we match the target and ground-truth using the highest intersection over union (IoU) among all decomposed elements. 
If no suitable match exists, the edit is not executed. 
Otherwise, we apply the same property change to the matched predicted element and render the edited prediction. 
We measure the structural similarity (SSIM)~\cite{ssim} between the ground-truth rendered image and the edited image. 
For text edits, we additionally evaluate whether the edited text is correctly realized by running OCR on the rendered result and reporting text recall against the ground truth rendered text. 

\begin{figure}[t!]
    \centering
    \includegraphics[width=1.0\linewidth]{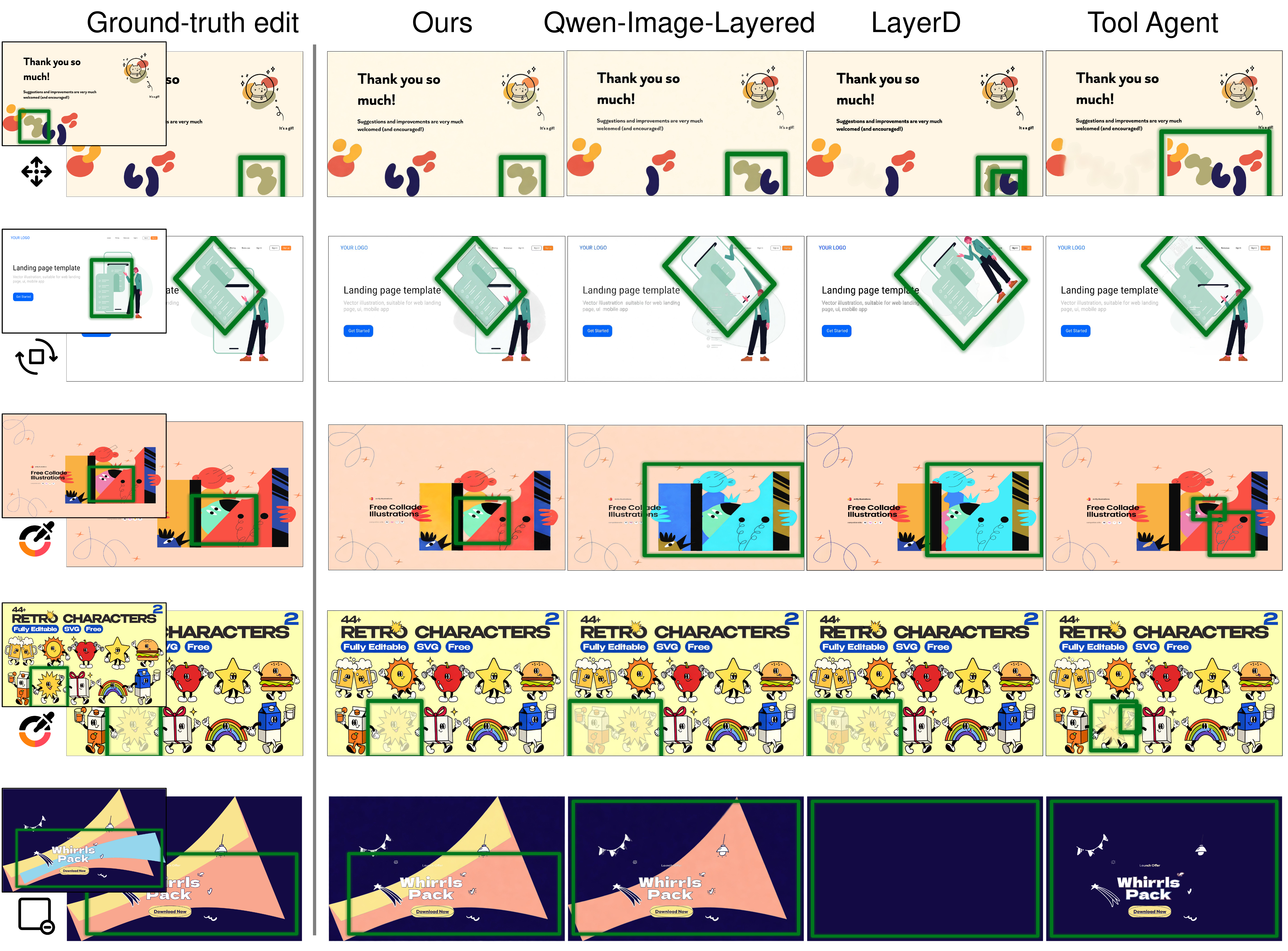}
    \vspace{-0.2cm}
    \caption{Example of edit instructions performed on editable formats produced by each baseline. Ground-truth edit pairs are the original image (top-left) and the edited image. Target edit areas are highlighted in green. Best viewed digitally.}
    \label{fig:exp-editability-quali}
    % \vspace{-1cm} %
\end{figure}

% Report the results (also mention the qualitative results)
\paragraph{\textup{\textbf{Editability results. }}}
As seen in \Cref{fig:exp-editability}, our method achieves the best editability across all edit types, showing that the recovered hierarchy supports both appearance edits and structural edits without breaking the design. 
In particular, we perform strongly on attribute edits such as recolor and opacity, where baselines often fail because the target element is not cleanly separated and the edit bleeds into background pixels or nearby layers. 
Qwen-Image-Layered~\cite{qwen-image-layered} and LayerD~\cite{layerd} remain competitive on simpler geometric edits such as delete and rotation, but their performance drops on edits that depend on correct element attribution and ordering, which requires the model to recover which pixels belong to which layer and how layers are stacked. 
We observe the same trend for text, where \Cref{fig:exp-editability}~(b) shows that our method achieves the highest text recall after replaying text edits, while the Tool Agent baseline degrades sharply because error cascades in long serial tool chains often corrupt the reconstruction early, and text is typically the first content to become distorted and undetectable in later steps.

Qualitative examples in \Cref{fig:exp-editability-quali} illustrate this failure mode, where edits intended for one element unintentionally affect several other elements. 
Overall, these results indicate that structured reconstruction yields representations that can be precisely targeted, and that graceful verification prevents early decomposition mistakes from propagating into failures for edits that require correct grouping and z-order.

\begin{table}[t]
\centering
\caption{Reconstruction accuracy measured on the Figma dataset. }
\vspace{-0.2cm}
\begin{tabular}{lcx{1.4cm}x{1.4cm}x{1.3cm}x{1.3cm}}
\toprule
\multirow{2}{*}{\textbf{Method}} &
\multicolumn{1}{c}{\textbf{Object-level}} &
\multicolumn{2}{c}{\textbf{Global-level}} &
\multicolumn{2}{c}{\textbf{Layout}} \\
\cmidrule(lr){2-2}\cmidrule(lr){3-4}\cmidrule(lr){5-6}
& \textbf{L1$\downarrow$} &
\textbf{PSNR$\uparrow$} &
\textbf{LPIPS$\downarrow$} &
\textbf{PQ$\uparrow$} &
\textbf{F1$\uparrow$} \\
\midrule
VTracer~\cite{vtracer} & 0.0977 & 20.487 & 0.1917 & 24.64 & 0.309 \\
LayerD~\cite{layerd} & 0.0704 & 16.141 & 0.3381 & 30.09 & 0.350 \\
Qwen-Image-Layered~\cite{qwen-image-layered} & 0.0493 & 26.192 & 0.1073 & 35.37 & 0.429 \\
Tool Agent~\cite{yao2022react} & 0.0493 & 13.923 & 0.3869 & 45.33 & 0.527 \\
\midrule
Ours & \textbf{0.0431} & \textbf{26.286} & \textbf{0.0883} & \textbf{45.37} & \textbf{0.535} \\
\bottomrule
\end{tabular}
\label{tab:exp-figma-accuracy}
% \vspace{-0.2cm}
\end{table}

\begin{table}[t]
\centering
\caption{Reconstruction accuracy measured on the Crello dataset~\cite{crello} }
\vspace{-0.2cm}
\begin{tabular}{lcx{1.4cm}x{1.4cm}x{1.3cm}x{1.3cm}}
\toprule
\multirow{2}{*}{\textbf{Method}} &
\multicolumn{1}{c}{\textbf{Object-level}} &
\multicolumn{2}{c}{\textbf{Global-level}} &
\multicolumn{2}{c}{\textbf{Layout}} \\
\cmidrule(lr){2-2}\cmidrule(lr){3-4}\cmidrule(lr){5-6}
& \textbf{L1$\downarrow$} &
\textbf{PSNR$\uparrow$} &
\textbf{LPIPS$\downarrow$} &
\textbf{PQ$\uparrow$} &
\textbf{F1$\uparrow$} \\
\midrule
VTracer~\cite{vtracer} & 0.0953 & 18.805 & 0.2919 & 19.55 & 0.243 \\
LayerD~\cite{layerd} & 0.0938 & 14.452 & 0.4585 & 30.98 & 0.369 \\
Qwen-Image-Layered~\cite{qwen-image-layered} & 0.0506 & \textbf{26.419} & \textbf{0.0985} & 40.19 & 0.496 \\
Tool Agent~\cite{yao2022react} & 0.0767 & 12.011 & 0.5674 & 44.16 & 0.527 \\
\midrule
Ours & \textbf{0.0465} & 23.525 & 0.1249 & \textbf{49.57} & \textbf{0.587} \\
\bottomrule
\end{tabular}
\label{tab:exp-crello-accuracy}
\end{table}

\begin{figure}[t!]
    \centering
    \includegraphics[width=1.0\linewidth]{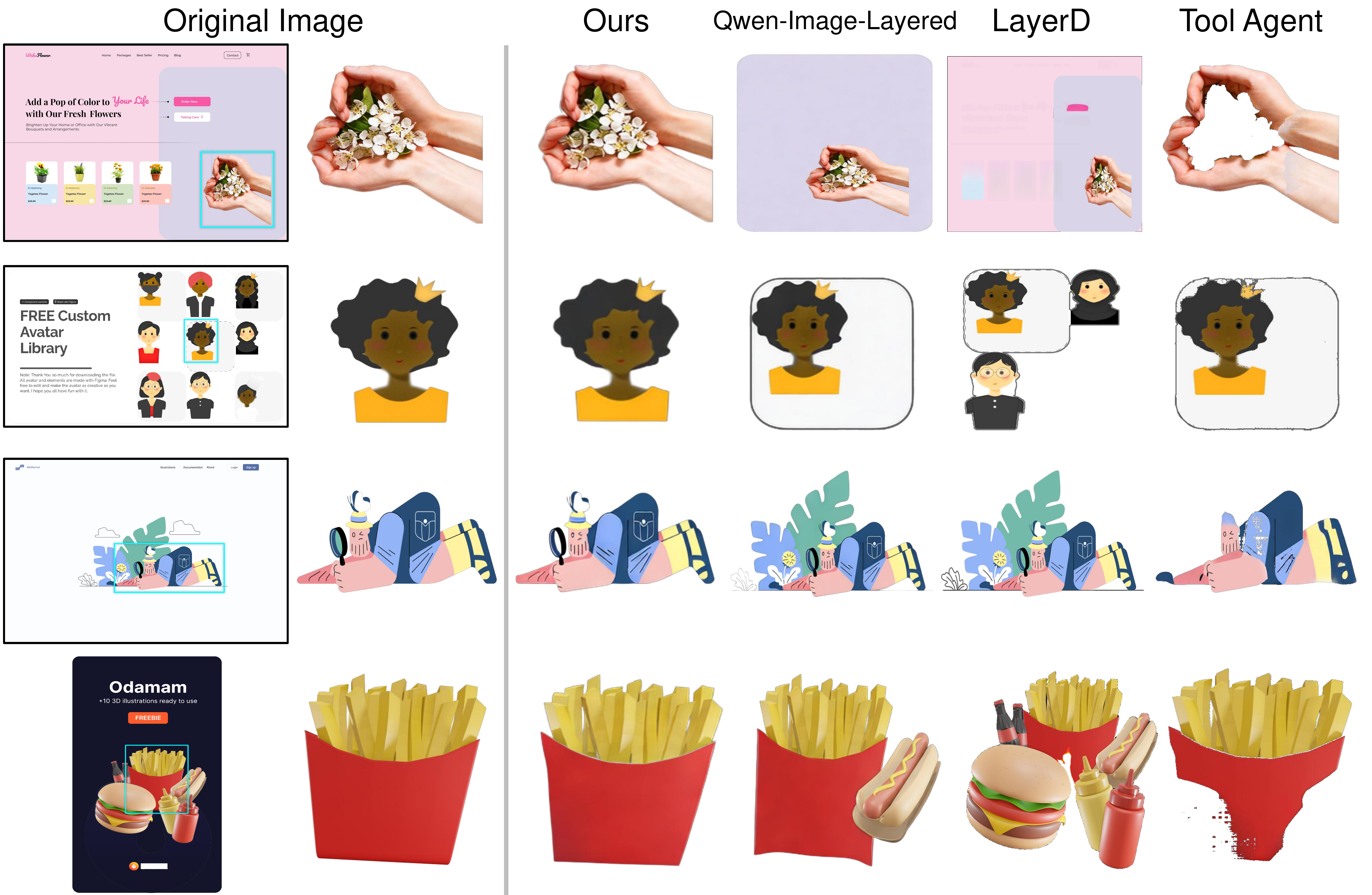}
    \caption{Decomposition accuracy comparison against baseline approaches. Previous approaches often fail to decompose multiple objects with high accuracy. }
    \label{fig:exp-accuracy-quali}
    % \vspace{-1cm} 
\end{figure}

\subsection{Reconstruction Accuracy}
We evaluate visual reconstruction accuracy on both the Figma design dataset and the Crello dataset. 
For each input raster image, each method produces an editable output in our target format, which we render back to a raster image. 
We compare the rendered reconstruction to the input using complementary metrics that capture object-level appearance (mean L1 error), global visual fidelity (PSNR and LPIPS~\cite{lpips}), and layout fidelity (panoptic quality (PQ)~\cite{pq} and detection F1), reported in \Cref{tab:exp-figma-accuracy,tab:exp-crello-accuracy}.

On Figma, our method is best across all metrics, improving object appearance, global fidelity, and layout. 
The Tool Agent baseline shows much weaker global fidelity than other approaches, supporting our design choice of structured hierarchical decomposition with graceful verification to prevent error accumulation in long, linear tool chains. 
Qwen-Image-Layered~\cite{qwen-image-layered} attains strong global similarity but lags in layout, consistent with layer decompositions that match overall appearance while mis-segmenting or mis-ordering elements.
Qualitative results are presented in \Cref{fig:exp-accuracy-quali}, where baseline approaches often trade off between plausible hierarchical layouts and precise object segmentation, whereas our method preserves both, yielding decompositions that are structurally coherent and cleanly editable. 

On Crello, trends are similar. 
LayerD and Qwen-Image-Layered improve due to closer training distribution, but our approach still achieves the strongest layout fidelity while remaining competitive on object-level and global metrics, without explicit training on the Crello dataset.

\vspace{-0.1cm}
\section{Analysis}
\vspace{-0.1cm}
\subsection{Inference Cost, Variance, and Parallelism}
\begin{figure}
    \centering
    \includegraphics[width=1.0\linewidth]{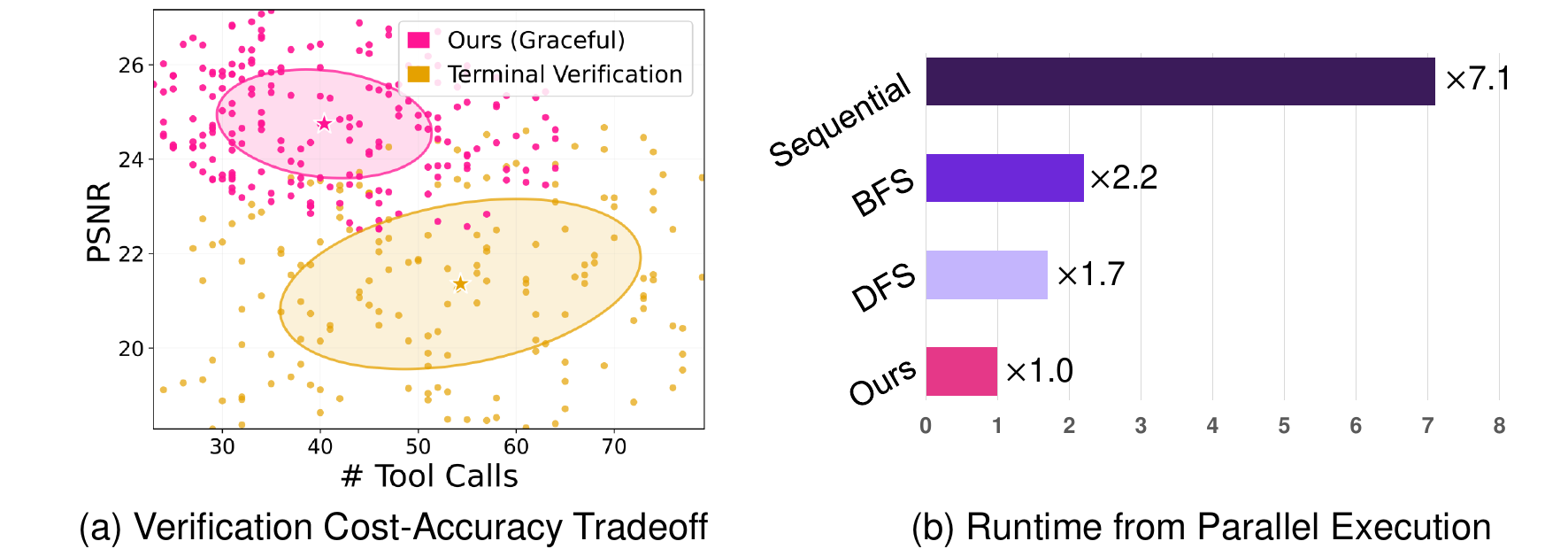}
    \caption{(a) Verification cost accuracy tradeoff. Graceful verification is faster, more accurate, and has lower variance than terminal verification. (b) Speedup from parallel tree expansion. Independent node expansions enable concurrent execution, yielding up to a $7.1\times$ speedup over serial tool use. }
    \label{fig:analysis-cost}
    \vspace{-0.2cm} 
\end{figure}

% Graceful verification is better than full verification
Agentic systems must account for cost, since each planning step and tool call consumes time and tokens, and one might expect that adding verification would only slow the pipeline down. 
In contrast, our results show that many small local verifications are cheaper than a few large terminal checks, because validating each structured decomposition step and repairing errors immediately avoids long error cascades that would otherwise force expensive restarts. 
In \Cref{fig:analysis-cost}(a), we plot time against accuracy (\ie, PSNR) for individual runs, and we find that our method is faster and more accurate than a sparse terminal verification, suggesting that keeping every step safe is often the shortest path through the search space. 
In addition, we find that our execution episodes have lower variance in the number of tool calls, indicating stable executions with minimal waste of tool calls.

% Right Parallel vs Serial
Furthermore, tool using agents are often slow for practical use because decisions and tool calls are typically executed in a single serial trajectory~\cite{li2025websailor,wu2025webdancer,li2025webweaver,wu2025webwalker}. 
In contrast, our reconstruction is organized as a tree, and each expansion depends only on its parent node and lineal history rather than on sibling states, so multiple frontier nodes can be expanded in parallel without concurrency conflicts. 
As a result, our approach runs $7.1$ times faster compared to a serial tool execution, as shown in \Cref{fig:analysis-cost}~(b). 
More broadly, our analysis reveals that formulating reconstruction as tree expansion enables speedups in both a technical sense, since independent nodes can be scheduled concurrently with minimal shared state, and a theoretical sense, since the dependency structure reduces the critical path length from the number of expansions to the depth of the hierarchy.

\begin{figure}
    \centering
    \includegraphics[width=\linewidth]{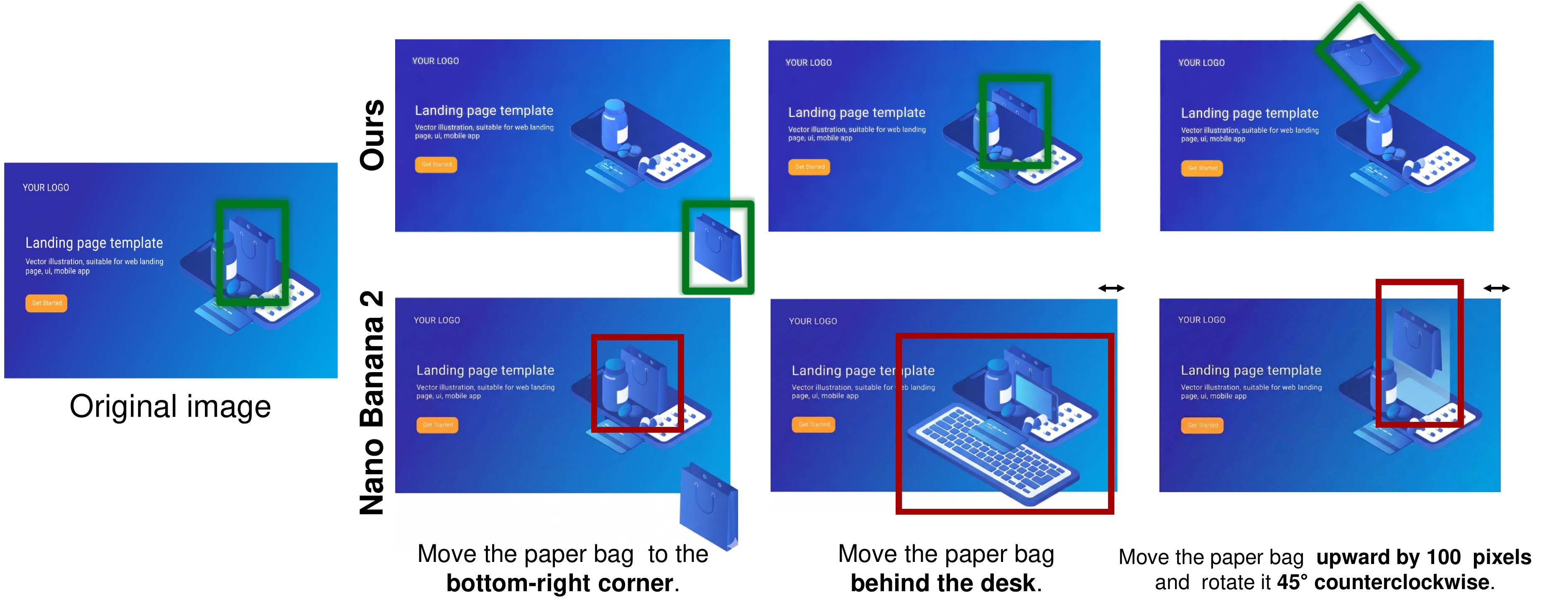}
    \vspace{-0.4cm}
    \caption{Nano Banana 2 comparison on spatial edits. For the same move and rotate instructions, Nano Banana 2 often changes unintended regions and change image sizes. }
    \label{fig:analysis-nanobanana}
    \vspace{-1.2cm}
\end{figure}

\subsection{Why Editable Formats Still Matter}
Image editing models~\cite{nanobanana,fluxkontext} are rapidly becoming more capable and can often produce visually plausible single step edits, which makes them appealing as a general solution for design iteration. 
Our comparison in \Cref{fig:analysis-nanobanana} shows that this flexibility does not necessarily indicate accuracy, as Nano Banana~2, a state-of-the-art image editing service, struggles with detailed spatial adjustments and also change aspect ratios even when the instruction is local. 
This highlights why recovering an explicit editable representation remains important, because a layer hierarchy with object identities and parameters supports deterministic layout changes, stable composition, and repeatable multi-step edits without unintended global warping.

\subsection{Limitations}

\begin{figure}[t]
    \centering
    \includegraphics[width=\linewidth]{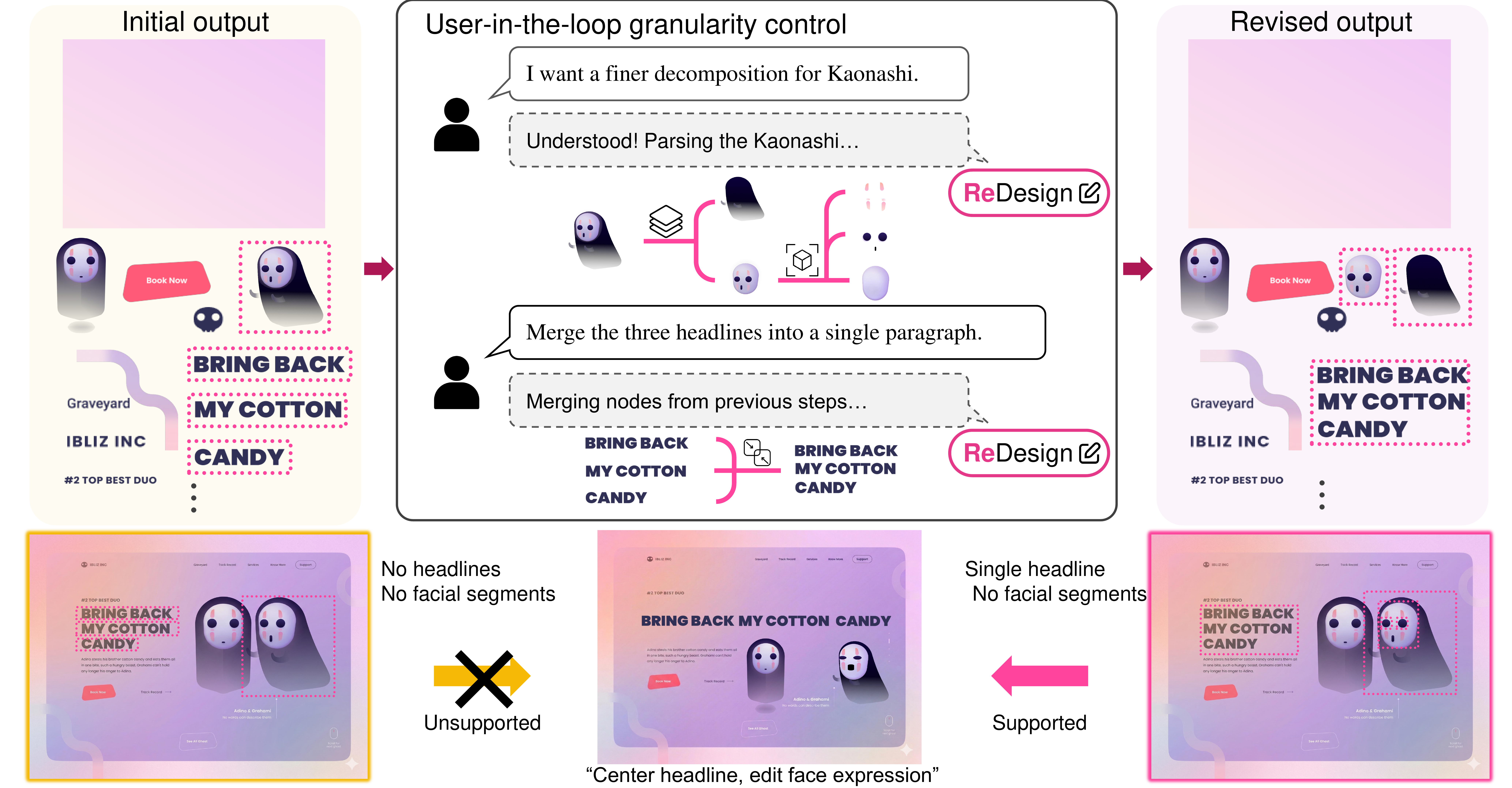}
    \vspace{-0.4cm}
    \caption{Our method offers the flexibility to adjust editing granularity through a user-in-the-loop revision.}
    \vspace{-0.5cm}
    \label{fig:analysis-limitation}
\end{figure}

% Something went wrong but we can always have controlability on fine-grained assets
% Ill-posedness of ~
A single notion of a good editable format is hard to define, since it depends on the intended edit and the level of control a designer needs. 
For example, vectorization can be unnecessary when the goal is a layout level adjustment, while fine grained paths matter when editing icons or shapes. 
Since we do not train to match the exact granularity of the Figma or Crello annotations, our reconstructions do not always align one to one with their layer splits, even when the rendered appearance is correct.

At the same time, not being tied to a fixed training distribution gives our system the flexibility to tune to any level of granularity. 
Since reconstruction is agentic, users can request a finer or coarser decomposition through prompting, and the system can continue expanding the tree to match the desired edit intent, as shown in \Cref{fig:analysis-limitation}. 
Moreover, the hierarchy naturally exposes higher level groups first, so complex vector paths can remain encapsulated and only be revealed when a user explicitly needs low level geometric control. 

\vspace{-0.2cm}
\section{Conclusion}
In this paper, we introduce an agentic framework that reconstructs editable design files from raster images by growing a layer hierarchy with specialized tools and enforcing graceful verification at every node expansion, so errors are detected and repaired locally instead of accumulating into hard failures. 
Across our new benchmark of raw Figma files with 14,796 edit replay instructions, ReDesign delivers strong visual fidelity while achieving the highest editability across layout, color, and text edits, and our analysis shows that step level verification and tree structured execution can be faster, more accurate, and more reliable than terminal checking and serial tool chains.

More broadly, our results suggest a practical direction for agentic systems, that imposing structural constraints on how an agent represents progress can simultaneously improve controllability, accuracy, and efficiency, because failure becomes local and repairable, and executions are predictable and thus scalable. 
In the perspective of designing workflows, despite the ever so growing capabilities of raster-based image editing, our method points to a future where editable representations remain essential, since they provide precise, real time control through explicit objects, parameters, and hierarchy, enabling users to make targeted changes with predictable behavior rather than relying on fragile pixel level edits. 

\section*{Acknowledgment}
This work was supported by Institute for Information \& communications Technology Planning \& Evaluation(IITP) grant funded by the Korea government(MSIT) (RS-2019-II190075, Artificial Intelligence Graduate School Program(KAIST)).
This work was supported by the National Research Foundation of Korea(NRF) grant funded by the Korea government(MSIT) (No. RS-2025-00555621).
This work was supported by the Technological Innovation R\&D Program (RS-2024-00438252) funded by the Ministry of SMEs and Startups(MSS, Korea). 

% ---- Bibliography ----
%
% BibTeX users should specify bibliography style 'splncs04'.
% References will then be sorted and formatted in the correct style.
%
% \clearpage
\bibliographystyle{splncs04}
\bibliography{main}

\input{appendix}

\end{document}

%% file: preamble.tex
\usepackage{graphicx}
\input{math_commands}

\usepackage{amssymb}
\usepackage{booktabs}

\usepackage{color}
\usepackage{varwidth}
\usepackage{xcolor}
\usepackage[table,xcdraw]{xcolor} % Table colors
\usepackage{soul} % Highlights and strike through

\usepackage{verbatim}
\usepackage{multirow}
\usepackage{adjustbox}
\usepackage{array}
\usepackage{bbm}
\usepackage{mathtools}
\usepackage{pifont}

\usepackage{scalerel}
\usepackage{listings}
\newcolumntype{x}[1]{>{\centering\let\newline\\\arraybackslash\hspace{0pt}}p{#1}}

\usepackage[symbol]{footmisc}

\usepackage{dirtytalk}
\usepackage{animate}
\usepackage{tocloft}
\usepackage{setspace}

\usepackage{lipsum}
\usepackage{soul}
\setuldepth{foobar}

%% file: math_commands.tex
\usepackage{amsmath,amsfonts,bm}

\def\eqref#1{equation~\ref{#1}}
\def\1{\bm{1}}

\DeclareMathAlphabet{\mathsfit}{\encodingdefault}{\sfdefault}{m}{sl}
\SetMathAlphabet{\mathsfit}{bold}{\encodingdefault}{\sfdefault}{bx}{n}

%% file: appendix.tex
\clearpage
\setcounter{section}{0}
\setcounter{page}{1}
\renewcommand*{\thesection}{\Alph{section}}
{\centering\Large\vspace{0.5em}
\textbf{Appendix} \\
\vspace{1.0em}
}

% Appendix
% Broader impact and ethical concerns (Pirating)
\section{Broader Impact and Ethical Considerations}

Recovering editable design structure from raster images can improve creative workflows by reducing the effort needed to recreate flattened assets such as screenshots or exports. 
This can make common tasks like revising text, colors, layout, accessibility properties, or localized content substantially easier, especially when the original source file is unavailable.

At the same time, this capability raises misuse concerns. 
Systems like ReDesign may lower the barrier to copying copyrighted graphics, branded assets, or commercial templates by making them easier to modify and reuse. 
We therefore view ReDesign as a tool for legitimate editing scenarios, such as revising authorized assets or restoring editability to one's own materials, rather than for bypassing ownership or attribution.
In deployment, such systems should be paired with safeguards, including clear communication that the recovered file is only an approximate reconstruction, usage policies around protected assets, and provenance or audit mechanisms that discourage uncredited replication.

\section{Extended Related Work}
\paragraph{\textbf{\textup{Layer-wise image generation.}}}
A growing body of work moves beyond flat RGB synthesis toward image representations composed of multiple transparent layers, motivated by the observation that layered structure provides a more editable intermediate representation than a single raster image. 
Early efforts explored foreground--background separation, limited transparent-layer synthesis, or post-hoc decomposition pipelines that first produce a flat image and then recover constituent regions or alpha layers~\cite{zhang2023text2layer,zhang2024transparent,huang2024layerdiff}. 
Recent methods scale this idea to richer RGBA decompositions by directly modeling interactions across layers, preserving transparency, and encouraging consistency between local elements and the final composite~\cite{fontanella2024generating,pu2025art,wang2025diffdecompose,kang2025layeringdiff,qwen-image-layered}. 
In parallel, another line of work studies decomposition rather than synthesis. 
Given a flattened image, the goal is to infer constituent layers, recover transparency, and separate visible elements so that the image becomes editable again~\cite{layerd,chen2025rethinking,omnipsd,nie2025decomposition}. 
Together, these works establish layered generation and decomposition as an important direction for controllable editing, while largely centering on learned end-to-end formulations for RGBA-style outputs.

\paragraph{\textbf{\textup{Vector graphics generation.}}}
A related direction studies converting raster images into vector graphics, which is especially useful for logos, icons, illustrations, and other shape-dominant assets. 
One family of methods formulates vectorization as optimization over differentiable rendering or iterative path fitting, seeking a compact vector program whose rendering matches the input image~\cite{diffvg,live,vectorfusion,svgdreamer,vtracer}. 
Another family treats SVG as a structured sequence and learns to generate vector graphics directly, improving semantic coherence and multimodal controllability over purely optimization-based formulations~\cite{deepsvg,starvector,omnisvg,layertracer}. 
Recent work further explores layered SVG generation, where vector primitives are organized into semantically meaningful groups rather than a single flat sequence~\cite{layertracer}. These methods are highly relevant to editability, but they primarily target vector-form reconstruction or generation. 
In realistic design recovery, only a subset of elements naturally admits vector representation, while text, raster imagery, and appearance-heavy effects are often better preserved in other forms. 
We use a forward-style vectorization model instead of an optimization based one due their improved speed and accuracy.

\section{Extended Analysis}

% Tree depth and action correlation.
\subsection{Action Selection Across Decomposition Tree Depth}

\begin{figure}
    \centering
    \includegraphics[width=\linewidth]{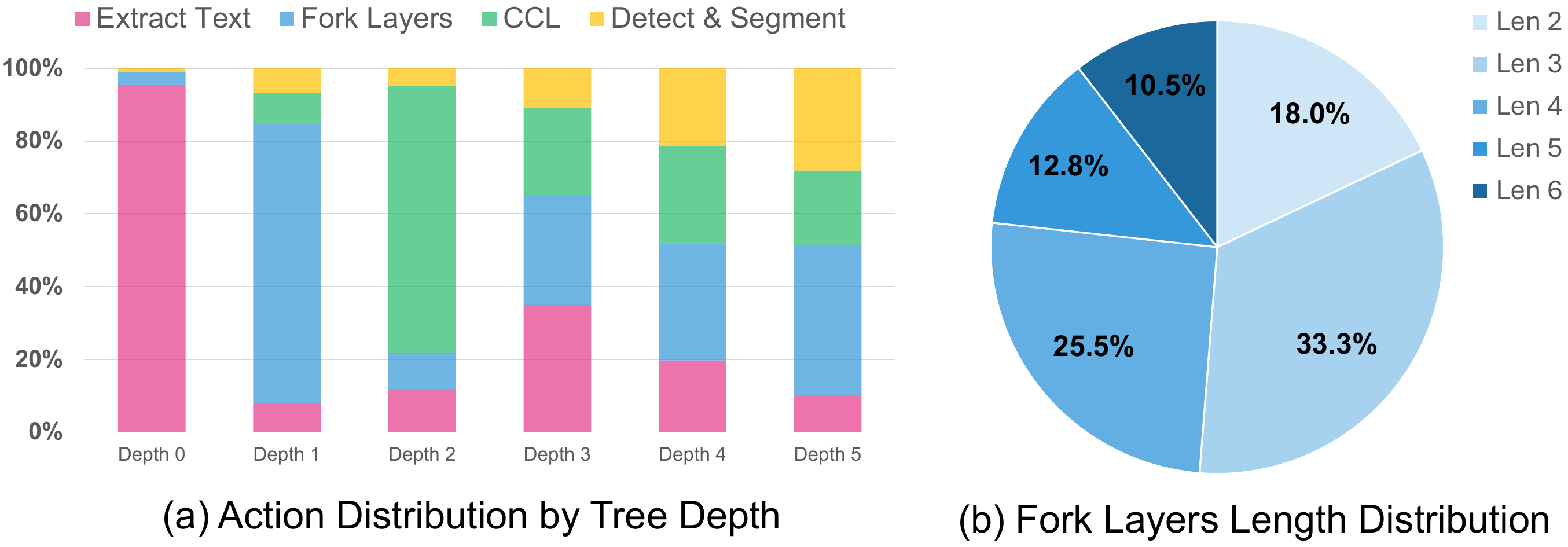}
    \vspace{-0.4cm}
    \caption{(a) Distribution of controller actions at each tree depth. (b) Layer-length hyperparameter distribution of Qwen-Image-Layered calls.}
    \label{fig:supple_depth_action_correlation}
\end{figure}

Figure~\ref{fig:supple_depth_action_correlation} reveals a clear \emph{coarse-to-fine decomposition strategy} that emerges in the controller.   
At shallow depths, the controller prioritizes \textit{Text Extraction}, reflecting an early commitment to preserving high-value, semantically precise elements such as text.  
At depth~1, it frequently invokes Qwen-Image-Layered to perform a broad structural split, producing coarse semantic layers that expose the major organization of the design.  
At depth~2, CCL becomes more common, as those coarse layers often decompose into spatially disconnected components that can be cleanly separated without generative inference.

Beyond these early stages, the controller is no longer guided by explicit depth-specific preferences and instead adapts its choices to the content of each node.  
In this regime, Detect \& Segment becomes increasingly prevalent, indicating that deeper nodes are dominated by localized, fine-grained refinements rather than global restructuring.  
This progression shows that the decomposition tree is not merely a bookkeeping device: it induces a meaningful division of labor across depth, with early levels handling semantically important and globally structured content, and later levels specializing in precise visual cleanup.

Figure~\ref{fig:supple_depth_action_correlation} (b) further shows that the controller adjusts the layer-length hyperparameter of Qwen-Image-Layered according to node complexity.  
Rather than applying a fixed decomposition granularity, the controller expands or restricts the number of proposed layers depending on how visually entangled the current node is.  
Together, these results suggest that our agent learns to use depth as an organizational prior, allocating different tools to different stages of decomposition in a way that is both structured and adaptive.

\begin{figure}
    \centering
    \includegraphics[width=0.6\linewidth]{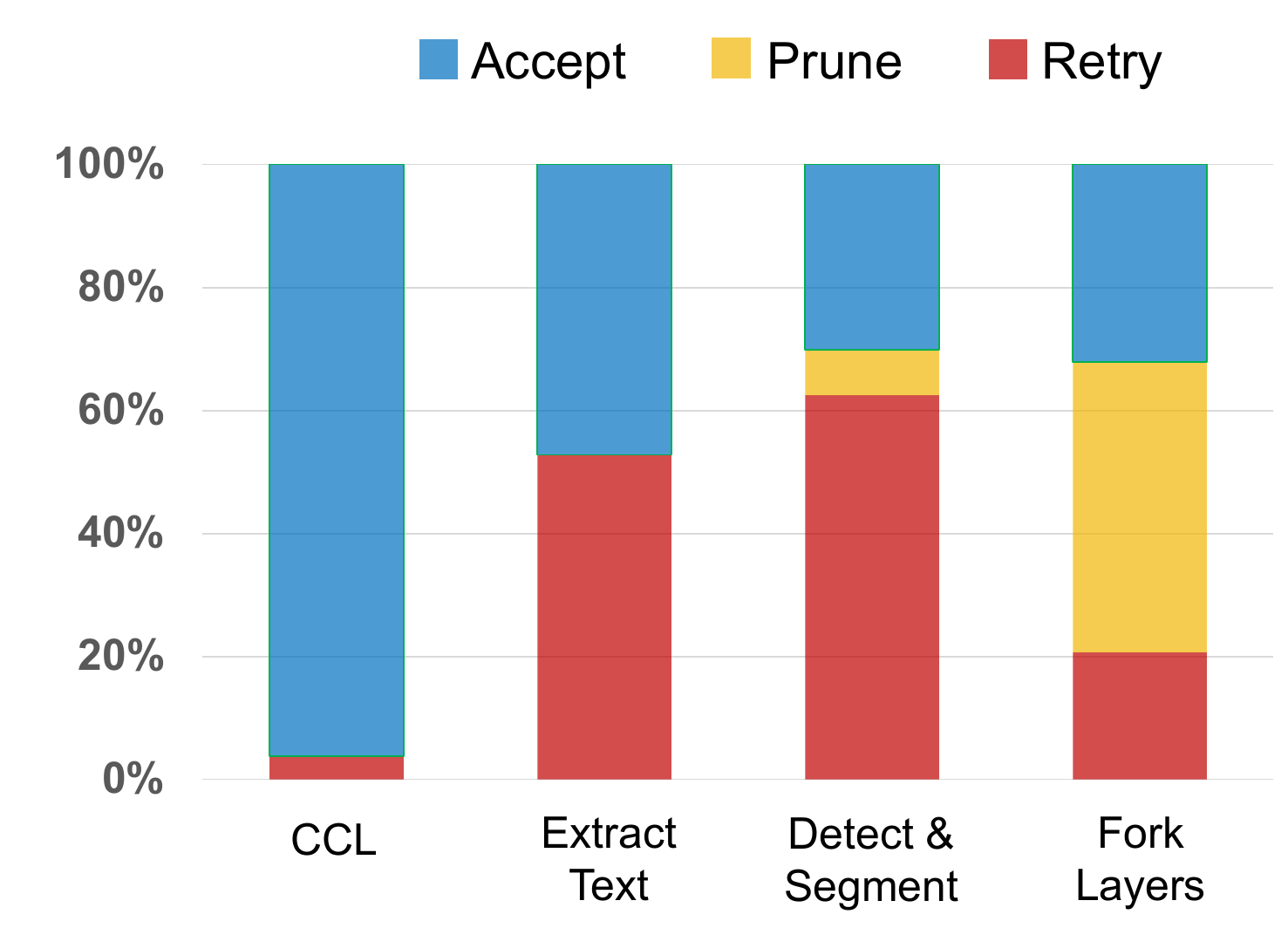}
    \caption{Distribution of verifier decisions per decomposition action.}
    \label{fig:supple_Verification_Analysis}
\end{figure}

% Graceful Verification Analysis
\subsection{Retry Rates during Graceful Verification}

\begin{figure}[t]
    \centering
    \includegraphics[width=1.0\linewidth, height=0.8\textheight, keepaspectratio]{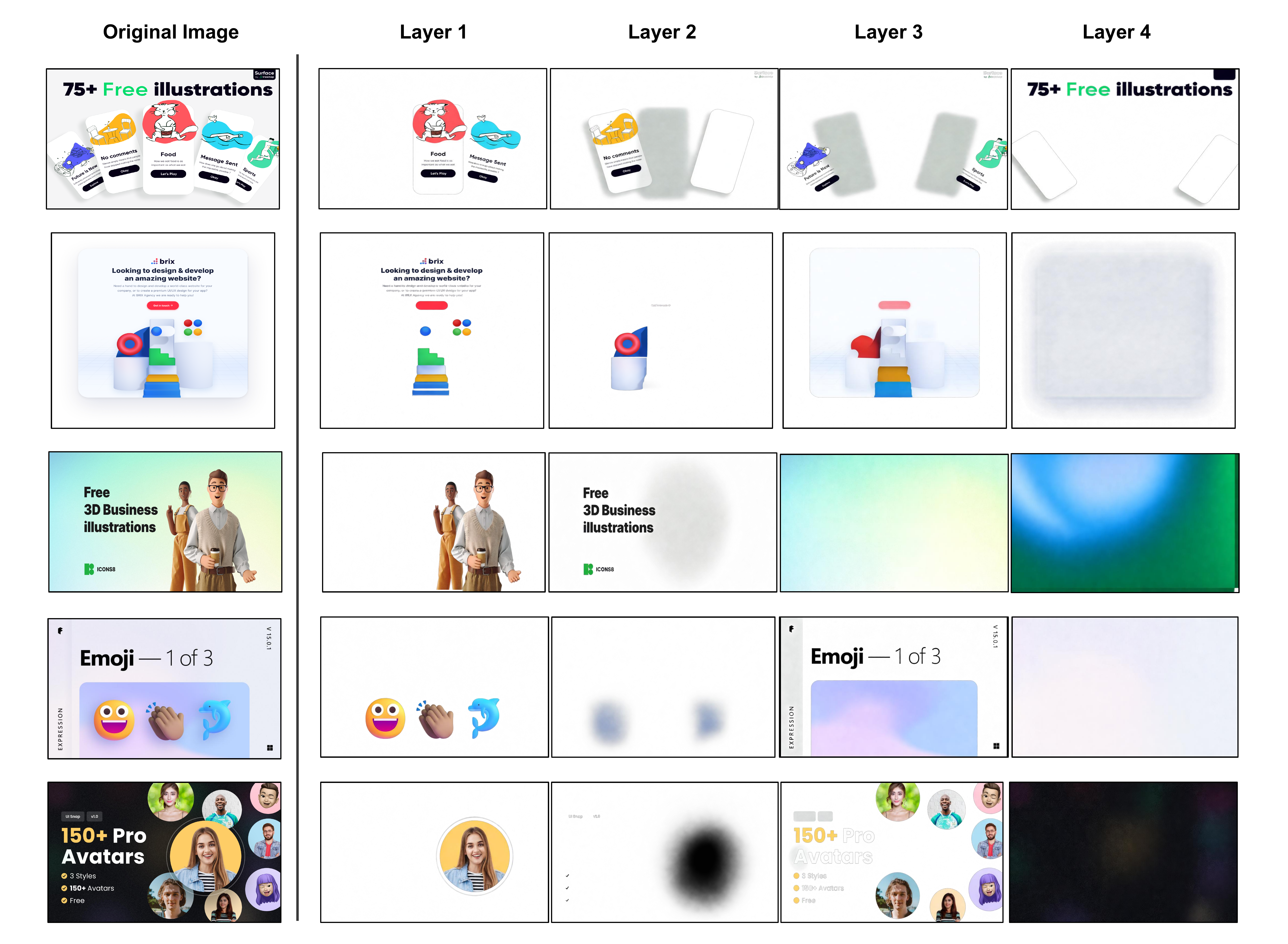}
    \caption{Failure cases of Qwen-Image-Layered on the Figma benchmark.}
    \label{fig:supple_qwen_failure_cases}
\end{figure}

\begin{figure}[t]
    \centering
    \includegraphics[width=1.0\linewidth, height=0.8\textheight, keepaspectratio]{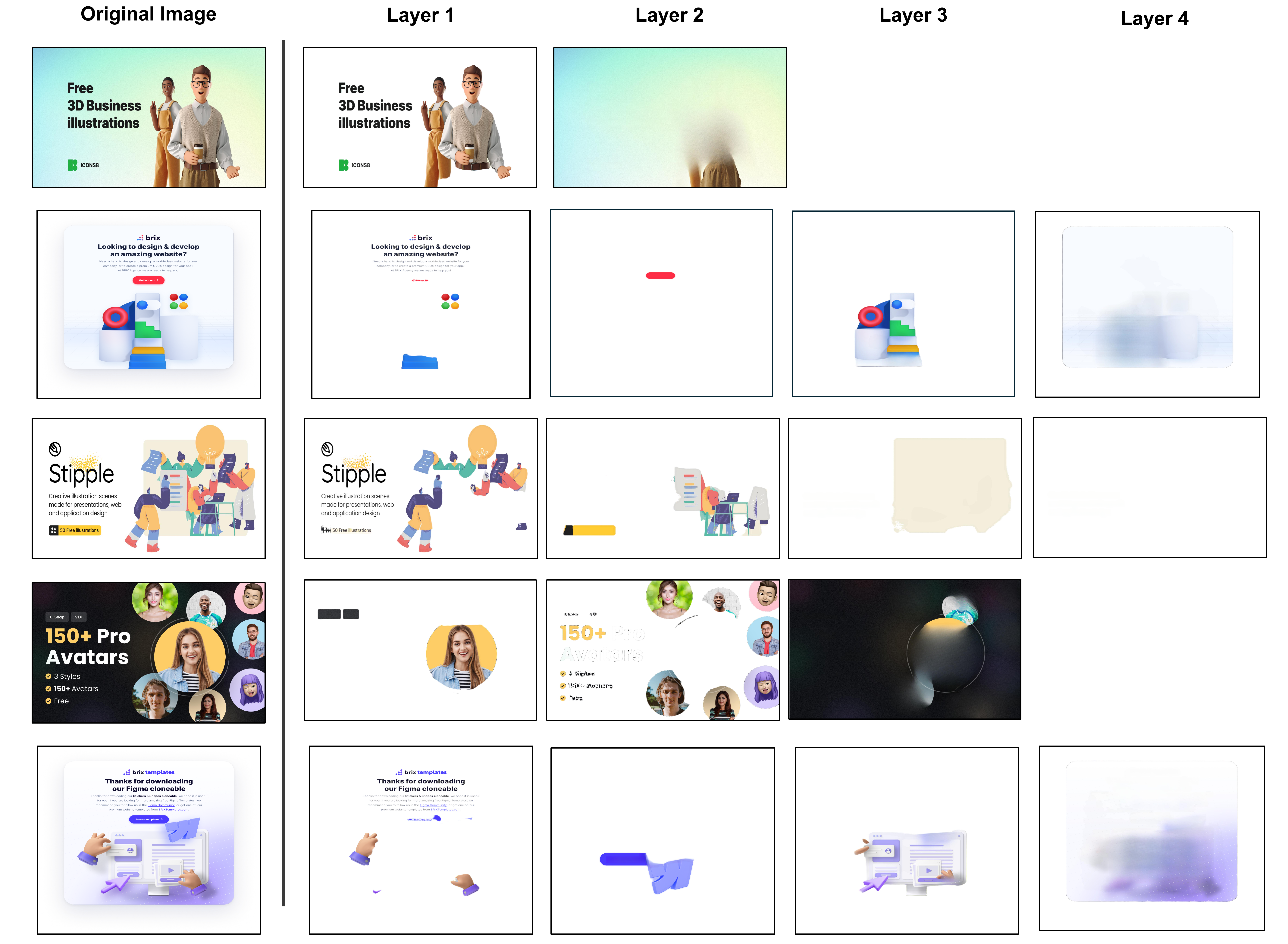}
    \caption{Failure cases of LayerD on the Figma benchmark.}
    \label{fig:supple_layerd_failure_cases}
\end{figure}

Figure~\ref{fig:supple_Verification_Analysis} shows the empirical
distribution of verifier outcomes across the four main decomposition actions.  
The verification behaviour is action-dependent by design. 
For CCL and Text Extraction, which directly extract content from the input, we restrict the verifier to binary accept-or-retry decisions since pruning would irreversibly discard information.  For Detect \& Segment and Fork Layers, which rely on generative models that occasionally hallucinate absent content, the verifier is additionally permitted to prune individual children. 
Except for CCL, a substantial fraction of tool outputs require
intervention through retry or partial pruning.
Figures~\ref{fig:supple_qwen_failure_cases} and~\ref{fig:supple_layerd_failure_cases} illustrate typical failure modes of Qwen-Image-Layered and LayerD, including hallucinated content, incomplete separations, and blurred artifacts, confirming that graceful verification is essential for filtering out erroneous outputs before they propagate into the reconstruction tree.

\raggedbottom
\subsection{Robustness to VLM Backbones}
\begin{table}[t]
\centering
\vspace{1.0cm}
\resizebox{0.8\linewidth}{!}{%
\begin{tabular}{lcx{1.4cm}x{1.4cm}x{1.3cm}x{1.3cm}}
\toprule
\multirow{2}{*}{\textbf{Controller / Verifier}} &
\multicolumn{1}{c}{\textbf{Object-level}} &
\multicolumn{2}{c}{\textbf{Global-level}} &
\multicolumn{2}{c}{\textbf{Layout}} \\
\cmidrule(lr){2-2}\cmidrule(lr){3-4}\cmidrule(lr){5-6}
& \textbf{L1$\downarrow$} &
\textbf{PSNR$\uparrow$} &
\textbf{LPIPS$\downarrow$} &
\textbf{PQ$\uparrow$} &
\textbf{F1$\uparrow$} \\
\midrule
Gemini-3 flash (Default) & \textbf{0.0431} & 26.286 & 0.088 & \textbf{45.37} & \textbf{0.530} \\
GPT-5 mini & 0.0517 & \textbf{26.930} & \textbf{0.079} & 43.39 & 0.521 \\
\bottomrule
\end{tabular}
}
\vspace{0.4cm}
\caption{Ablation on the VLM used for the controller and verifier. Gemini corresponds to our default configuration reported in Table~\ref{tab:exp-figma-accuracy}.}
\label{tab:ablation-controller-llm}
\end{table}

\begin{figure*}
    \centering
    \includegraphics[width=0.65\linewidth]{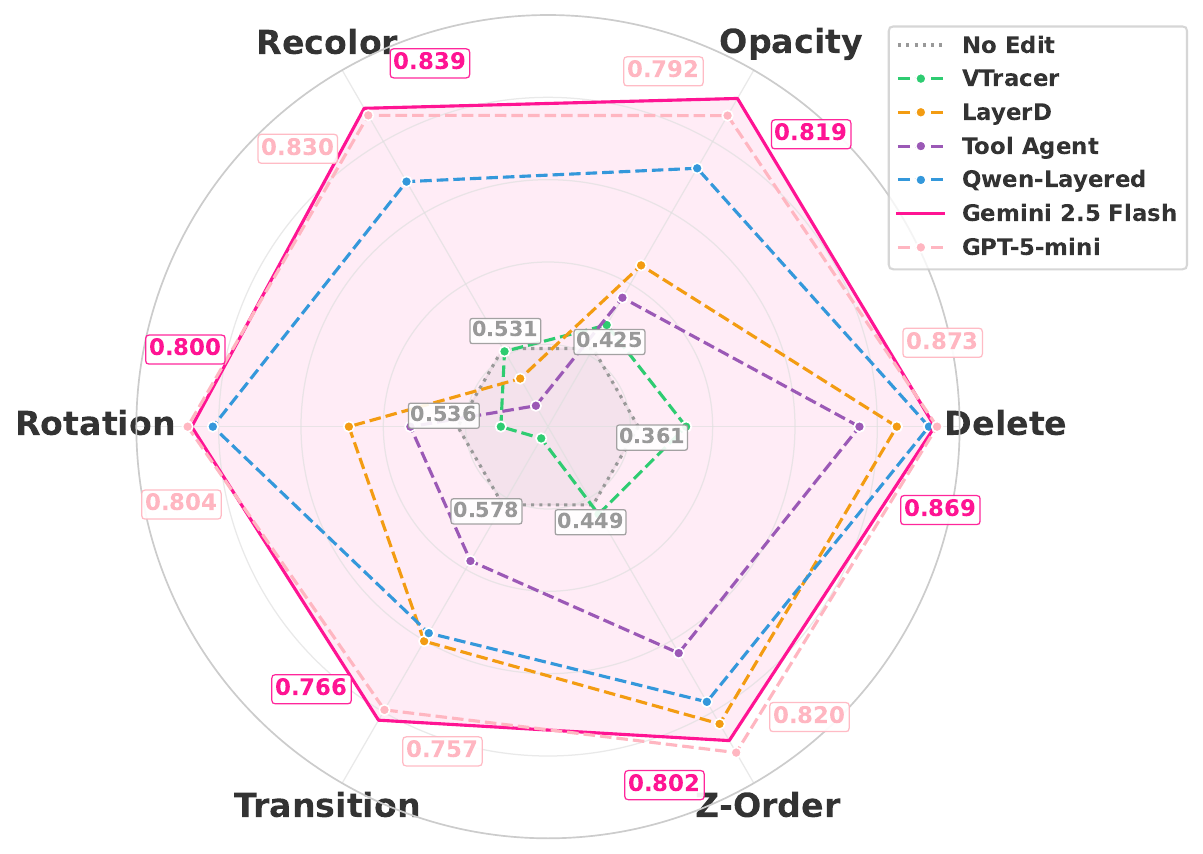}
    \label{fig:supp-gpt-editability}
    \caption{Robustness to VLM backbones. Using a different VLM (GPT-5-mini) show similar results to our default setting, both achieving state-of-the-art performance in creating editable design decompositions.}
\end{figure*}

We compare two VLMs as the controller and verifier, Gemini-3-flash (our default) and GPT-5 mini in \Cref{tab:ablation-controller-llm}. 
The two models slightly exhibit a coverage-fidelity trade-off governed by verification strictness.
GPT-5 mini applies stricter verification, selectively retaining only high-confidence elements. 
This improves global-level fidelity (PSNR and LPIPS) by keeping the final composite free of less certain reconstructions.
Gemini analyzes multi-image context more thoroughly and avoids over-pruning, resulting in stronger element-level metrics (L1, PQ, and F1). 
The broader element coverage slightly shifts the composite distribution, but better preserves the full structure of the original design. 
Nonetheless, the difference between these two VLM backbones is marginal, and our structured workflow enables robust performance gains compared to a naive tool using agent. 
We adopt Gemini as our default, as higher element coverage better serves downstream editability.

\begin{table}[t]
\vspace{-0.5cm}
\centering
\caption{Additional ablations on a subset of Figma. }
\label{tab:sup-ablation}
\resizebox{0.8\linewidth}{!}{%
\begin{tabular}{lccccccc}
\toprule
\multirow{2}{*}{Method}
& \multicolumn{1}{c}{Object-level}
& \multicolumn{2}{c}{Global-level}
& \multicolumn{2}{c}{Layout}
& \multicolumn{2}{c}{Cost} \\
\cmidrule(lr){2-2} \cmidrule(lr){3-4} \cmidrule(lr){5-6} \cmidrule(lr){7-8}
& L1$\downarrow$ & PSNR$\uparrow$ & LPIPS$\downarrow$ & PQ$\uparrow$ & F1$\uparrow$ & Tool Calls$\downarrow$ & API Cost$\downarrow$ \\
\midrule
No Memory               & 0.0542 & 25.36 & 0.0985 & 39.09 & 0.4604 & 31.4 & \$0.032 \\
Sequential execution        & 0.0621 & 25.22 & 0.0964 & 42.00 & 0.4951 & 36.8 & \$0.040 \\
\midrule
Tree depth = 2          & 0.0550 & \textbf{28.43} & \textbf{0.0645} & 31.92 & 0.381 & \textbf{6.9} & \textbf{\$0.013} \\
Tree depth = 3          & 0.0594 & 27.10 & 0.0871 & 41.77 & 0.4939 & 12.0 & \$0.020 \\
Tree depth = 4          & 0.0541 & 26.06 & 0.0946 & 42.79 & 0.5034 & 26.7 & \$0.025 \\
\midrule
Ours                    & \textbf{0.0474} & 25.56 & 0.0969 & \textbf{44.76} & \textbf{0.5194} & 28.8 & \$0.027 \\
\bottomrule
\end{tabular}
}
\vspace{-0.3cm}
\end{table}

\subsection{Quantitative ablation study}
We report quantitative results in \Cref{tab:sup-ablation} on experiment carried out by varying the tree depth and memory hierarchy. 
The results align with our prior analysis, where shallow depth produces coarse decompositions (high accuracy and less editability) while full depth gives us higher layout accuracy and thus higher editability.

\section {Implementation Details}
% How we match elements during the edit tasks
\subsection{Edit Replay Protocol}
\label{sec:supp_edit_replay}

We formulate the matching between GT and predicted elements as a
minimum-cost many-to-many bipartite assignment. Let $\mathcal{G}=\{g_1,\dots,g_M\}$ and
$\mathcal{P}=\{p_1,\dots,p_N\}$ denote the GT and predicted element
sets, where each element~$e$ carries an RGBA image
$\mathbf{I}_e\!\in\![0,1]^{H\times W\times 4}$ and a binary mask
$\mathbf{m}_e\!\in\!\{0,1\}^{H\times W}$ on a shared canvas.
Because elements may occlude one another, we compute a \emph{visible
mask} for each element in descending $z$-order.

\begin{equation}
  \mathbf{v}_k
    = \mathbf{m}_k \;\wedge\; \lnot\,\mathbf{O}_k,
  \qquad
  \mathbf{O}_k
    = \bigvee_{j:\,z_j>z_k} \mathbf{m}_j\,.
  \label{eq:visible_mask}
\end{equation}

To handle many-to-many correspondences arising from over- or under-segmentation relative to the GT design structure, we
generate candidate groups via directional containment ratios. \begin{equation}
  c^{\text{gt}}_{i\to j}
    = \frac{|\mathbf{v}_i\wedge\mathbf{v}_j|}
           {|\mathbf{v}_j|+\epsilon}\,,
  \qquad
  c^{\text{pred}}_{j\to i}
    = \frac{|\mathbf{v}_i\wedge\mathbf{v}_j|}
           {|\mathbf{v}_i|+\epsilon}\,.
  \label{eq:containment}
\end{equation}
For each GT element~$g_i$, predicted elements with
$c^{\text{gt}}_{i\to j}\!\ge\!\tau_h$ form a merge candidate
set~$\mathcal{M}_i$, from which multiple subsets are enumerated as candidate match targets, allowing the matcher to recognise valid parses even when the predicted decomposition differs from the GT design structure.  
The procedure runs symmetrically for each $p_j$ to build GT-side groups~$\mathbb{G}_\text{gt}$.

For each group pair $(G_i,P_j)$ we composite members in $z$-order and compute a matching cost.
\begin{equation}
  \mathcal{C}(G_i,P_j)
    = \lambda_{\ell_1}\,\ell_1
    + \lambda_\text{IoU}(1-\text{IoU})
    + |G_i|\cdot\rho_\text{gt} 
    + |P_j|\cdot\rho_\text{pred}
  \label{eq:cost}
\end{equation}
where $\ell_1$ is the mean absolute RGB difference over the union of
visible masks and IoU is computed on binary masks.  The merge penalties
$\rho_\text{gt}$ and $\rho_\text{pred}$ prevent the solver from
aggressively merging unrelated elements to achieve superficially low
costs, ensuring that matching respects the original design structure.

Given the cost matrix, we solve for a globally optimal matching via the Hungarian algorithm.

\begin{equation}
  \min_{\mathbf{x}}\;
    \textstyle\sum_{i,j}\mathcal{C}(G_i,P_j)\,x_{ij}
  + \tau_d \, n_{\text{unmatched}}\,,
  \label{eq:assignment}
\end{equation}

where $x_{ij}\!\in\!\{0,1\}$ indicates whether group pair $(G_i,P_j)$ is selected.  The cost matrix is padded with dummy entries of cost~$\tau_d$, so an element is left unmatched only when every real candidate exceeds~$\tau_d$.

Based on the matched pairs, we evaluate reconstruction accuracy and editability.  For
editability, we apply six edits to both GT and predicted sets.
\textit{Delete} removes a matched element by zeroing all RGBA channels.
\textit{Opacity} shifts the alpha channel by $\Delta\alpha\in[-1,\,1]$.
\textit{Recolor} transforms element pixels in HSV space with hue
shift $\Delta_H$, saturation multiplier~$\mu_S$, and brightness
multiplier~$\mu_V$.
\textit{Rotation} rotates an element by ~$\theta\in[-180^{\circ},\,180^{\circ}]$.
\textit{Translation} displaces an element by $(d_x,d_y)$, within the canvas area.
\textit{Z-order} exchanges the stacking index of an element with its nearest overlapping neighbor.
After each edit, we re-composite the full canvas for both GT and
predicted element sets and measure the discrepancy within the edited region.

% Implementation details
\subsection{System Configuration}

\begin{table}[t]
\centering
\small
\setlength{\tabcolsep}{20pt}
\begin{tabular}{@{}ll@{}}
\toprule
\textbf{Tool / Library} & \textbf{Purpose} \\
\midrule
Grounding DINO (Swin-B)
  & Open-vocab.\ object detection \\[3pt]
PaddleOCR v5
  & Text detection \& recognition \\[3pt]
HiSAM
  & Text region segmentation \\[3pt]
SAM 2.1 Hiera-L
  & Box-prompted segmentation \\[3pt]
LaMa (Big)
  & Mask-based inpainting \\[3pt]
ObjectClear
  & Diffusion-based removal \\[3pt]
SciPy (\texttt{ndimage.label})
  & Connected Component Labeling \\[3pt]
VTracer
  & Raster $\to$ SVG vectorisation \\[3pt]
Qwen-Image-Layered
  & Multi layer decomposition \\[3pt]
WhatFontIs API
  & Font family prediction \\
\bottomrule
\end{tabular}
\vspace{10pt}
\caption{Tool configurations. All tools are used with their default settings, without any additional fine-tuning.}
\label{tab:tools}
\end{table}

Our system is a training-free agentic pipeline built on the publicly available tools detailed in \Cref{tab:tools} and commercial VLM APIs. 
The pipeline is orchestrated as a state machine using
LangGraph~\cite{langgraph}.  
For our experiments, we use a NVIDIA A6000 to locally host tools that require acceleration (\eg, Qwen-Image-Layered~\cite{qwen-image-layered}, detection, segmentation, and inpainting). 
With optimization libraries~\cite{yin2026vllm}, the tools we use can run on commercial GPUs under 8GB of VRAM. 

\begin{figure}[t]
    \centering
    \includegraphics[width=1.0\linewidth]{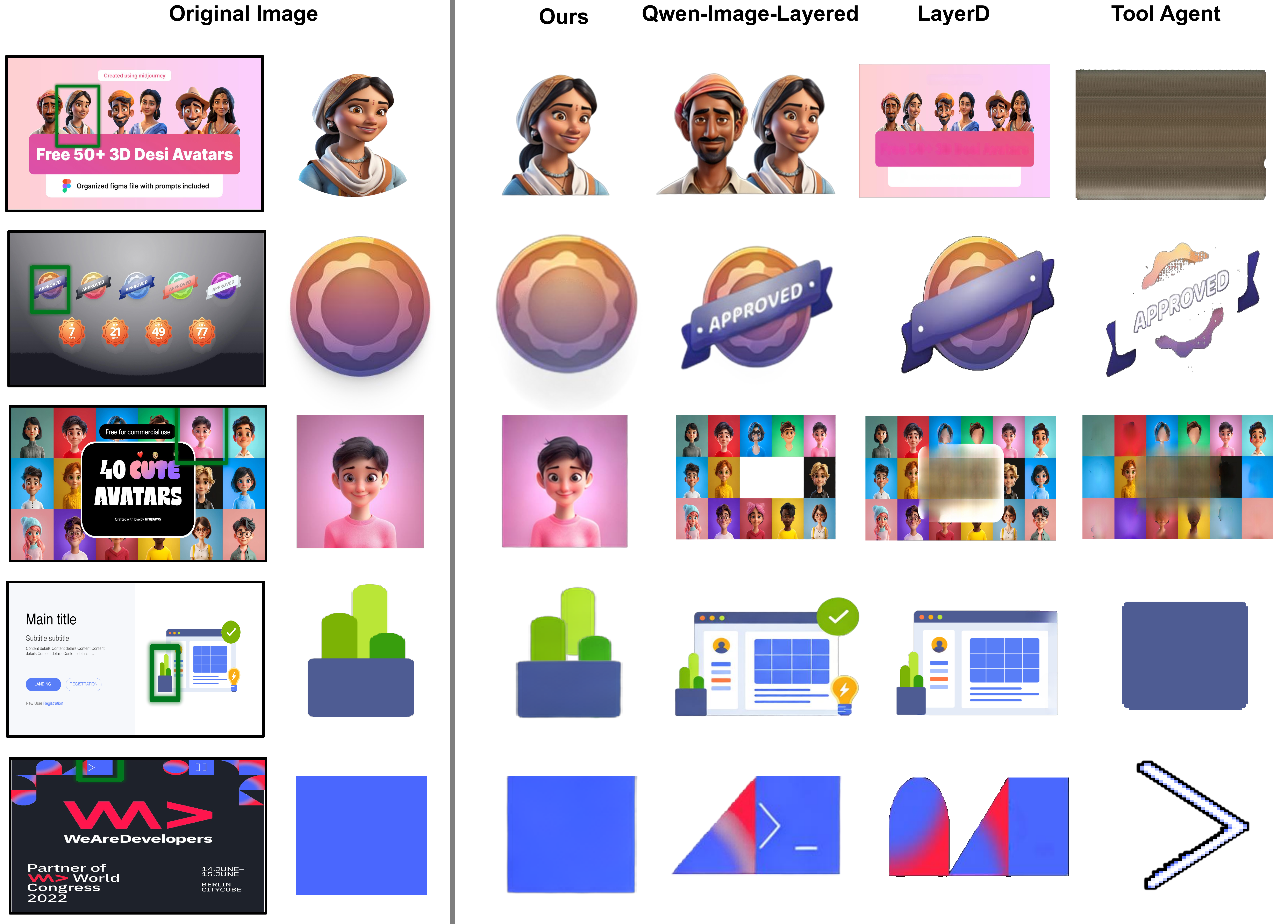}
    \vspace{-0.5cm}
    \caption{Additional decomposition accuracy comparison against baseline approaches. Best viewed digitally.}
    \label{fig:supple_extra_accuracy_images}
    % \vspace{-1cm} 
\end{figure}

\begin{figure}[b]
    \centering
    \includegraphics[width=1.0\linewidth, height=0.85\textheight, keepaspectratio]{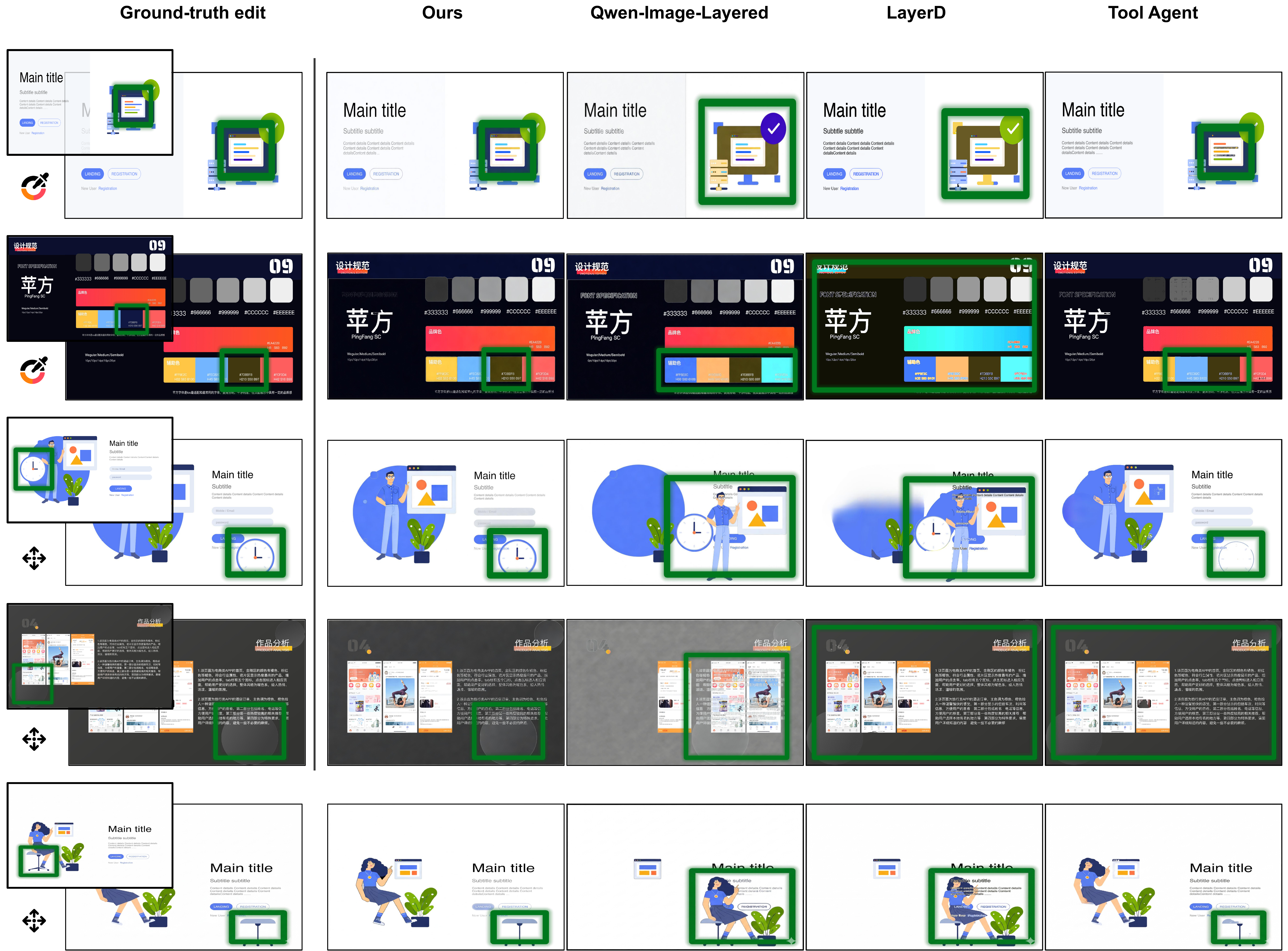}
    \vspace{-0.5cm}
    \caption{Additional edit replay results compared against baseline approaches. Best viewed digitally.}
    \label{fig:supple_extra_edit_images}
    % \vspace{-1cm} %
\end{figure}

% More results
\subsection{Additional Qualitative results}
We present additional qualitative comparisons on the Figma benchmark in \Cref{fig:supple_extra_accuracy_images,fig:supple_extra_edit_images}. 
Our method recovers clean element boundaries and correct stacking order even under heavy occlusion, allowing edits to remain well-localised to the target element. 
This benefit extends to both attribute changes and spatial transformations.

% Prompts
\subsection{Prompts}
\begin{figure}[t!]
    \centering
    \includegraphics[width=1.0\linewidth]{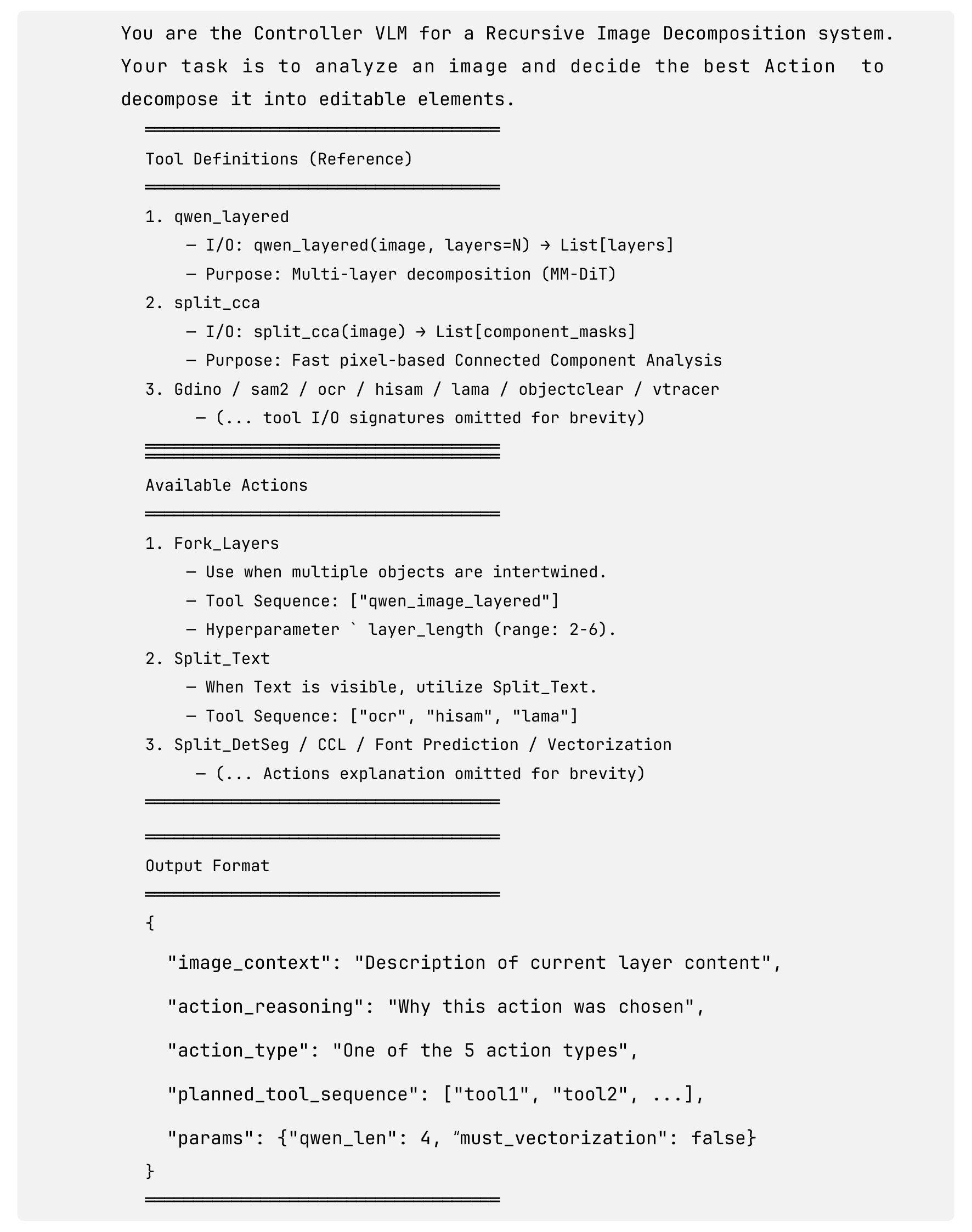}
    \vspace{-0.5cm}
    \caption{System prompt used for the VLM controller.}
    \label{fig:supple_prompt1}
    % \vspace{-1cm} %
\end{figure}

\begin{figure}[t!]
    \centering
    \includegraphics[width=1.0\linewidth]{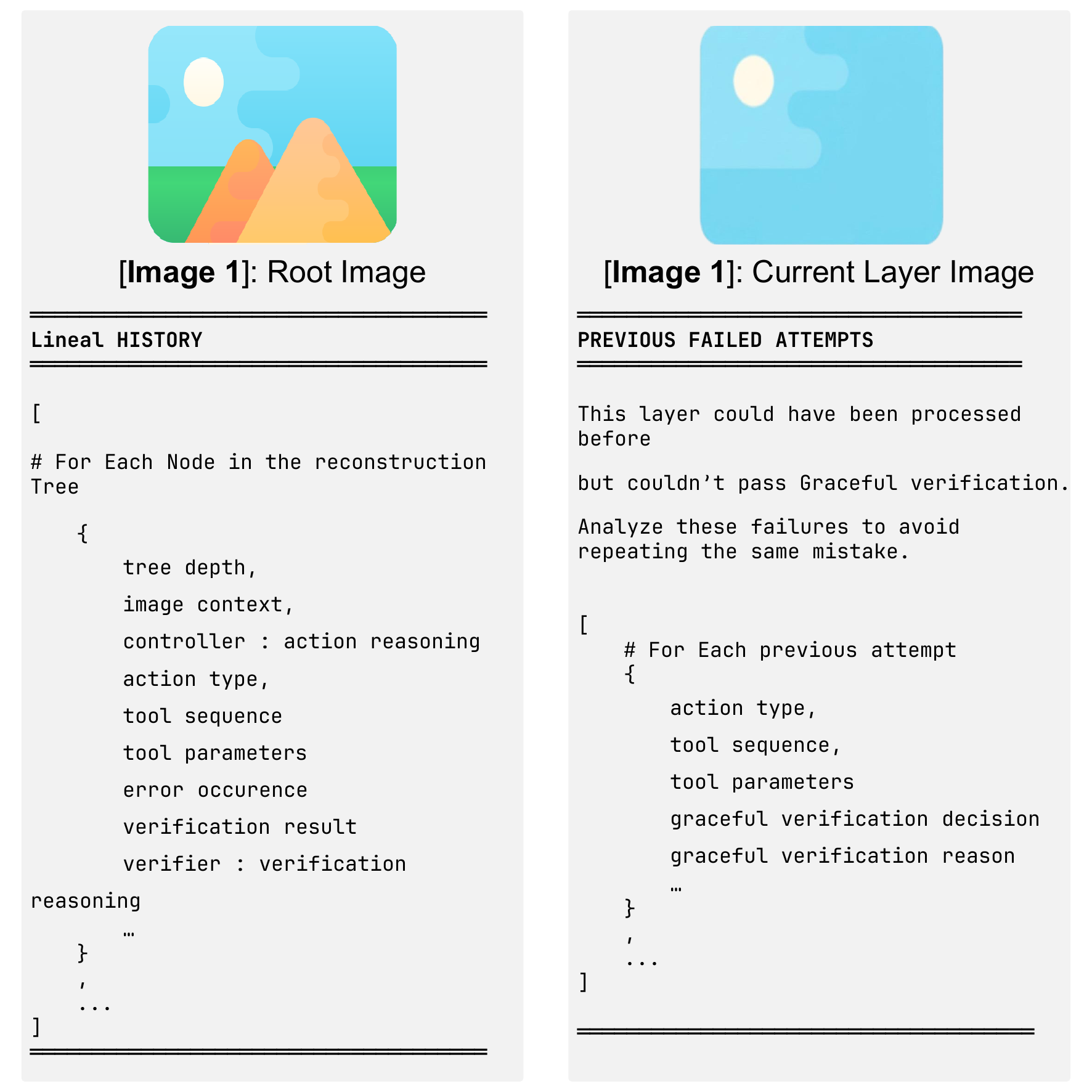}
    \vspace{-0.5cm}
    \caption{Linear history and failure history templates.}
    \label{fig:supple_prompt2}
    % \vspace{-1cm} %
\end{figure}

We provide the prompts used by the two VLM agents in our framework.
As shown in \Cref{fig:supple_prompt1,fig:supple_prompt2}, the Controller system prompt defines the available tools, action types, and expected JSON output schema.
For each node, the Controller additionally receives a node-specific prompt containing the root image, the current layer image, the lineal history of prior decomposition decisions, and any previous failed attempts with their verification feedback, enabling context-aware planning.

\begin{figure}[t!]
    \centering
    \includegraphics[width=1.0\linewidth]{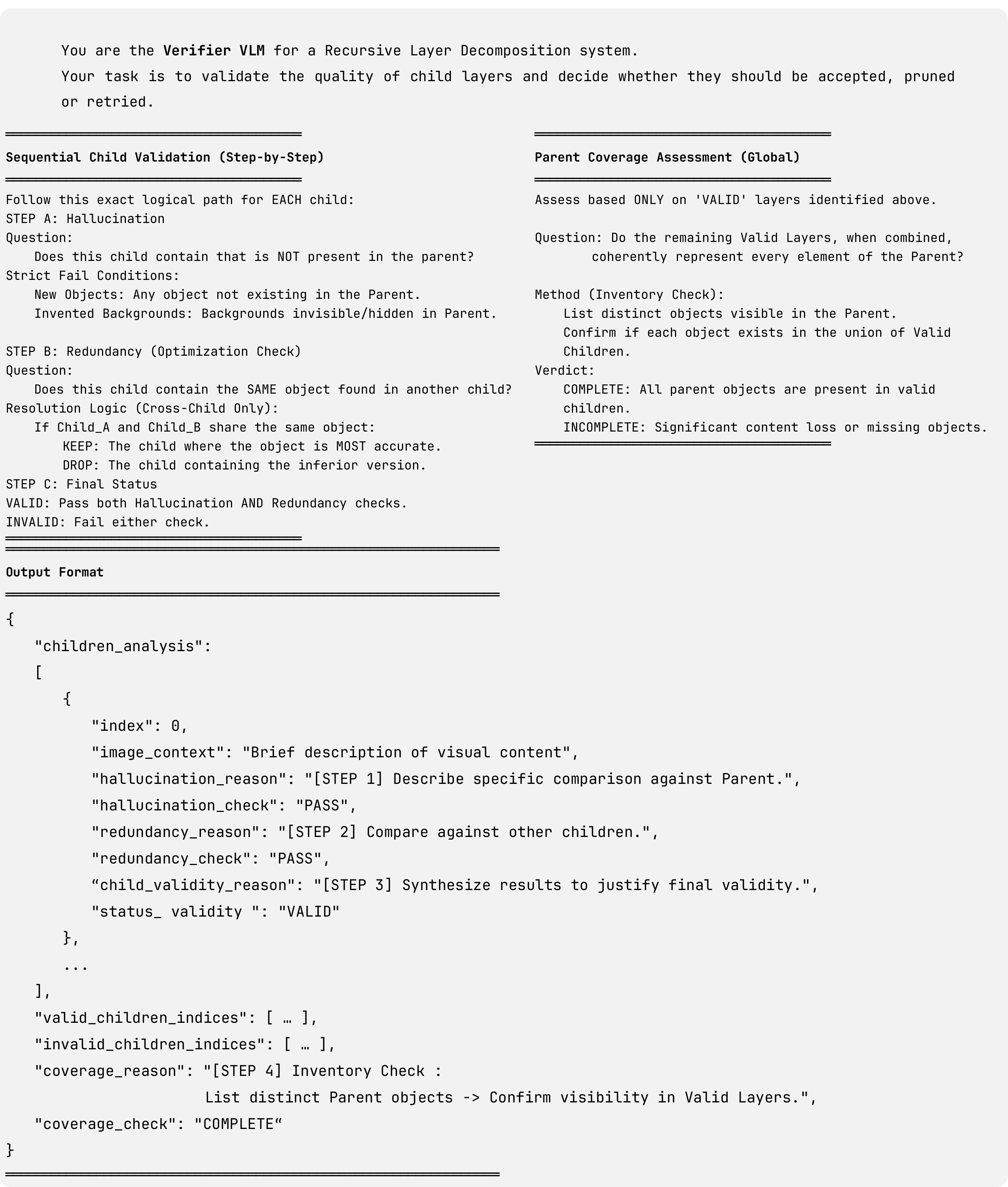}
    \vspace{0.5cm}
    \caption{System prompt used for graceful verification.}
    \label{fig:supple_prompt3}
    % \vspace{-1cm} %
\end{figure}

The Verifier system prompt in \Cref{fig:supple_prompt3} specifies a three-step sequential validation protocol of hallucination checking against the parent, cross-child redundancy detection, and parent coverage assessment that confirms whether the union of valid children accounts for all visual content in the parent. 

% \begin{figure}[t!]
%     \centering
%     \includegraphics[width=0.8\linewidth]{figures/supple_prompts4.pdf}
%     \vspace{0.5cm}
%     \caption{Tool calling definition}
%     \label{fig:supple_prompt4}
%     % \vspace{-1cm} %
% \end{figure}